\pdfoutput=1

\documentclass[11pt]{article}
\usepackage[dvipsnames, table]{xcolor}
\usepackage[most]{tcolorbox}
\usepackage{authblk}
\usepackage[]{coling}

\usepackage{times}
\usepackage{latexsym}

\usepackage[T1]{fontenc}

\usepackage[utf8]{inputenc}

\usepackage{microtype}

\usepackage{lipsum}
\usepackage{graphicx}
\usepackage{hyperref}
\usepackage{algorithm}
\usepackage{algorithmic}
\usepackage{nicematrix,booktabs}
\usepackage{multirow}
\usepackage{ragged2e}
\usepackage{tabularx}
\usepackage{tcolorbox}
\usepackage{subcaption}
\usepackage{transparent}
\usepackage{listings}
\usepackage{multicol}
\usepackage{boldline}
\usepackage{arydshln}
\usepackage{diagbox}
\usepackage{amssymb}
\usepackage{pifont}
\newcommand{\cmark}{\ding{51}}
\newcommand{\xmark}{\ding{55}}
\usepackage{tikz}
\usepackage{enumitem}
\usepackage{changepage}

\definecolor{LightCyan}{RGB}{232,241,255}
\definecolor{LightRed}{RGB}{255,235,235}
\definecolor{LightPink}{RGB}{255,235,255}
\definecolor{LightGreen}{RGB}{218,255,234}
\definecolor{LightYellow}{RGB}{255,255,235}
\definecolor{LightGray}{RGB}{242,242,242}
\definecolor{Red}{RGB}{253, 239, 242}
\definecolor{Yellow}{RGB}{255, 255, 204}
\definecolor{Pink}{RGB}{255, 243, 254}
\definecolor{Gray}{RGB}{249, 249, 249}
\definecolor{Green}{RGB}{230, 255, 241}
\definecolor{Blue1}{RGB}{218, 232, 245}
\definecolor{Blue2}{RGB}{239, 248, 253}
\definecolor{Blue3}{RGB}{136, 190, 220}
\definecolor{Blue4}{RGB}{83, 157, 204}
\definecolor{Blue5}{RGB}{42, 122, 185}
\definecolor{Blue6}{RGB}{11, 85, 159}
\definecolor{GreenCheck}{RGB}{0, 102, 51}
\definecolor{LightBack}{RGB}{247,249,251}

\tcbset{
    guideline/.style={
    colback=gray!5!white, 
        colframe=white, 
        boxrule=0.5mm, 
        left=1mm, 
        right=1mm, 
        top=1mm, 
        bottom=1mm, 
        arc=3mm, 
        boxsep=3mm,
        drop shadow={black!50!white}, 
        enhanced,
        overlay={
            \node[fill=cyan!70!black, text=white, rounded corners=1mm, font=\bfseries\scriptsize, inner sep=1mm] 
            at (frame.north west) 
            [xshift=8mm, yshift=-0mm] {Guidelines};
            }
    }
}

\tcbset{
    negation/.style={
        colback=gray!5!white, 
        colframe=white, 
        boxrule=0.5mm, 
        left=1mm, 
        right=1mm, 
        top=1mm, 
        bottom=1mm, 
        arc=3mm, 
        boxsep=3mm,
        drop shadow={black!50!white}, 
        enhanced,
        overlay={
            \node[fill=purple!70!white, text=white, rounded corners=1mm, font=\bfseries\scriptsize, inner sep=1mm] 
            at (frame.north west) 
            [xshift=8mm, yshift=-0mm] {Negation};
        }
    }
}

\tcbset{
    prompt/.style={
        colback=gray!5!white, 
        colframe=white, 
        boxrule=0.5mm, 
        left=1mm, 
        right=1mm, 
        top=1mm, 
        bottom=1mm, 
        arc=3mm, 
        boxsep=3mm,
        drop shadow={black!50!white}, 
        enhanced,
        overlay={
            \node[fill=blue!50!white, text=white, rounded corners=1mm, font=\bfseries\scriptsize, inner sep=1mm] 
            at (frame.north west) 
            [xshift=8mm, yshift=-0mm] {Prompt};
        }
    }
}

\tcbset{
    example/.style={
        colback=gray!5!white, 
        colframe=white, 
        boxrule=0.5mm, 
        left=1mm, 
        right=1mm, 
        top=1mm, 
        bottom=1mm, 
        arc=3mm, 
        boxsep=3mm,
        drop shadow={black!50!white}, 
        enhanced,
        overlay={
            \node[fill=red!50!white, text=white, rounded corners=1mm, font=\bfseries\scriptsize, inner sep=1mm] 
            at (frame.north west) 
            [xshift=8mm, yshift=-0mm] {Example};
        }
    }
}

\definecolor{backcolour}{rgb}{0.95,0.95,0.92}
\definecolor{codegreen}{rgb}{0,0.6,0}
\definecolor{mygreen}{HTML}{88EABB}
\definecolor{OliveGreen}{HTML}{00693E}
\definecolor{markgreen}{rgb}{0.3, 0.73, 0.09}
\definecolor{markred}{rgb}{0.8, 0.0, 0.0}

\lstdefinestyle{myStyle}{
    backgroundcolor=\color{backcolour},   
    commentstyle=\color{codegreen},
    basicstyle=\ttfamily\tiny,
    breakatwhitespace=false,         
    breaklines=true,                 
    keepspaces=true,                 
    numbers=none,       
    numbersep=5pt,                  
    showspaces=false,                
    showstringspaces=false,
    showtabs=false,                  
    tabsize=2,
    frame=single
}

\title{Piecing It All Together: Verifying Multi-Hop Multimodal Claims}

\author{\textbf{Haoran Wang$^{\heartsuit}$} \quad \textbf{Aman Rangapur$^{\heartsuit}$} \quad \textbf{Xiongxiao Xu$^{\heartsuit}$} \quad \textbf{Yueqing Liang$^{\heartsuit}$} \\ \quad \textbf{Haroon Gharwi$^{\heartsuit}$} \quad \textbf{Carl Yang$^{\clubsuit}$} \quad \textbf{Kai Shu$^{\clubsuit}$}\thanks{Corresponding author} \\
        {$\heartsuit$ Illinois Institute of Technology \quad $\clubsuit$ Emory University}\\
        \tt{\{hwang219, arangapur, xxu85, yliang40, hgharwi\}@hawk.iit.edu} \\
        \tt{\{j.carlyang, kai.shu\}@emory.edu} \\
}
\begin{document}
\maketitle

\begin{abstract}
Existing claim verification datasets often do not require systems to perform complex reasoning or effectively interpret multimodal evidence. To address this, we introduce a new task: multi-hop multimodal claim verification. This task challenges models to reason over multiple pieces of evidence from diverse sources, including text, images, and tables, and determine whether the combined multimodal evidence supports or refutes a given claim. To study this task, we construct MMCV, a large-scale dataset comprising 15k multi-hop claims paired with multimodal evidence, generated and refined using large language models, with additional input from human feedback. We show that MMCV is challenging even for the latest state-of-the-art multimodal large language models, especially as the number of reasoning hops increases. Additionally, we establish a human performance benchmark on a subset of MMCV. We hope this dataset and its evaluation task will encourage future research in multimodal multi-hop claim verification. Data and code are available: \href{https://mmcv-dataset.github.io/}{https://mmcv-dataset.github.io/}
\end{abstract}

\begin{table*}[!t]
\renewcommand{\arraystretch}{1.2}%
    \centering
    \setlength{\tabcolsep}{4pt}
    \resizebox{0.99\linewidth}{!}{%
        \begin{Tabular}{lccccc}
        \toprule
        
        \textbf{Dataset} & \textbf{Multimodal} & \textbf{Multi-hop} & \textbf{Evidence Retrieval} & \textbf{Annotated Evidence} & \textbf{Annotated Label} \\ \midrule
        
        \rowcolor{gray!10} FEVER \cite{thorne-etal-2018-fever}               & \textcolor{markred}{\xmark} & \textcolor{markred}{\xmark} & \textcolor{markgreen}{\cmark} & \textcolor{markgreen}{\cmark} & \textcolor{markgreen}{\cmark}     \\
        
        Liar \cite{wang-2017-liar}                        & \textcolor{markred}{\xmark} & \textcolor{markred}{\xmark} & \textcolor{markred}{\xmark} & \textcolor{markgreen}{\cmark} & \textcolor{markgreen}{\cmark}    \\
        
        \rowcolor{gray!10} FakeNewsNet \cite{shu2020fakenewsnet}             & \textcolor{markgreen}{\cmark} & \textcolor{markred}{\xmark} & \textcolor{markgreen}{\cmark} & \textcolor{markred}{\xmark} & \textcolor{markgreen}{\cmark}    \\
        
        NewsCLIPpings \cite{luo-etal-2021-newsclippings}  & \textcolor{markgreen}{\cmark} & \textcolor{markred}{\xmark} & \textcolor{markgreen}{\cmark} & \textcolor{markred}{\xmark} & \textcolor{markgreen}{\cmark}    \\
        
        \rowcolor{gray!10} Factify \cite{mishra2022factify}                  & \textcolor{markgreen}{\cmark} & \textcolor{markred}{\xmark} & \textcolor{markred}{\xmark} & \textcolor{markred}{\xmark} & \textcolor{markred}{\xmark}    \\
        
        COSMOS \cite{aneja2021cosmos}                     & \textcolor{markgreen}{\cmark} & \textcolor{markred}{\xmark} & \textcolor{markgreen}{\cmark} & \textcolor{markred}{\xmark} & \textcolor{markgreen}{\cmark}    \\
        
        \rowcolor{gray!10} InfoSurgeon \cite{fung-etal-2021-infosurgeon}     & \textcolor{markgreen}{\cmark} & \textcolor{markred}{\xmark} & \textcolor{markgreen}{\cmark} & \textcolor{markred}{\xmark} & \textcolor{markgreen}{\cmark}    \\
        
        Fauxtography \cite{zlatkova-etal-2019-fact}       & \textcolor{markgreen}{\cmark} & \textcolor{markred}{\xmark} & \textcolor{markred}{\xmark} & \textcolor{markred}{\xmark} & \textcolor{markgreen}{\cmark}    \\
        
        \rowcolor{gray!10} HoVer \cite{jiang-etal-2020-hover}                & \textcolor{markred}{\xmark} & \textcolor{markgreen}{\cmark} & \textcolor{markgreen}{\cmark} & \textcolor{markgreen}{\cmark} & \textcolor{markgreen}{\cmark}    \\
        
        Mocheg \cite{yao2023end}                          & \textcolor{markgreen}{\cmark} & \textcolor{markred}{\xmark} & \textcolor{markgreen}{\cmark} & \textcolor{markgreen}{\cmark} & \textcolor{markgreen}{\cmark}    \\
        \hline
        
        \rowcolor{gray!10} \textbf{MMCV (Ours)} & \textcolor{markgreen}{\textbf{\cmark}} & \textcolor{markgreen}{\textbf{\cmark}} & \textcolor{markgreen}{\textbf{\cmark}} & \textcolor{markgreen}{\textbf{\cmark}} & \textcolor{markgreen}{\textbf{\cmark}}     \\ \bottomrule
        \end{Tabular}
    }
\captionsetup{width=0.99\linewidth}
\caption{Comparison between \textsc{MMCV} and other claim verification datasets. The columns indicate whether the dataset requires multimodal content, multi-hop reasoning, explanation generation, and whether it contains annotated evidence.}
\label{tab:comparison}
\end{table*}

\section{Introduction}
\label{sec:introduction}
Due to the rapid growth in AI-generated content \cite{huang2024authorship, huang2024can, zhang2024llm, jin2024agentreview}, it is difficult for automated fact-checking systems to keep up with verifying the accuracy of claims with multimodal evidence. This challenge is further exacerbated by the recent development of diffusion models such as DALL-E \cite{ramesh2021zero} and Stable Diffusion \cite{rombach2022high}, which can generate realistic images from textual prompts \cite{liu2024sora}. These powerful tools could enable attackers to produce misleading information \cite{10.1145/3627673.3679821, pan-etal-2023-risk, huang2024social, gao2024best, jin2024mm} at a low cost.  Additionally, these claims often require multi-hop reasoning, where a set of connected evidence pieces leads to the final verdict of a claim \cite{yang-etal-2018-hotpotqa}. As a result, there is a need for automated tools to assist human fact-checkers in evaluating the veracity of multimodal multi-hop claims.

\begin{figure}[!t]
    \centering
    \includegraphics[width=\linewidth]{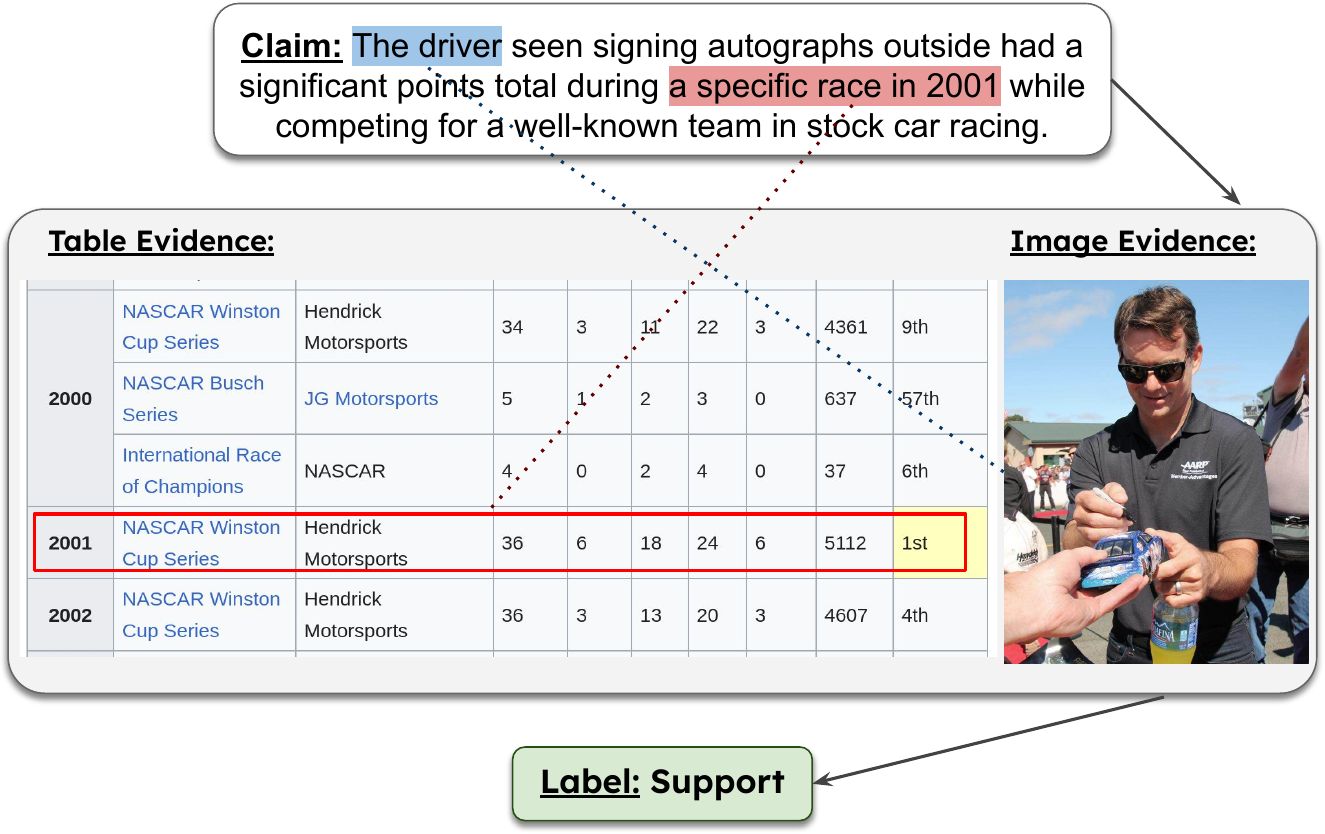}
    \caption{An illustration of a 2-hop claim from \textsc{MMCV}. To correctly verify this claim, the system must reason over both the image evidence and the table evidence.}
    \label{fig:example}
\end{figure}

Claim verification, which involves assessing the veracity of an input claim against a collection of evidence, is a vital tool in combating the spread of misinformation \cite{thorne-vlachos-2018-automated, guo-etal-2022-survey, jin2022towards, jin2023predicting, yang2022reinforcement}. However, verifying multi-hop multimodal claims introduces new challenges in both dataset construction and effective modeling. Unlike single-hop claims, which require only straightforward one-step reasoning, multi-hop claims require multiple reasoning steps to reach a final verdict. Furthermore, the inclusion of multimodal evidence requires models to understand and integrate information across various modalities, such as text, images, and tables, making it more complex to comprehend and extract relevant information. For instance, to verify the claim shown in Figure \ref{fig:example}, a system must understand the semantic content of the image, integrate all relevant information from the table evidence, and apply multi-step reasoning to arrive at the final conclusion.

In this paper, we introduce the task of multi-hop multimodal claim verification to evaluate the veracity of multi-hop claims against multimodal evidence. To study this task, we construct \textbf{\underline{M}}ulti-hop \textbf{\underline{M}}ultimodal \textbf{\underline{C}}laim-\textbf{\underline{V}}erification \textbf{\textsc{(MMCV)}}, a dataset of 15K multi-hop claims paired with multimodal evidence that either \textsc{SUPPORT} or \textsc{REFUTE} each claim. To create the dataset, we develop a novel pipeline that uses large language models (LMMs) for data annotation, supported by human feedback. This method significantly reduces the workload on human annotators and cuts costs, while ensuring high quality and factual accuracy of the dataset. Our pipeline first uses LLMs to re-formulate multi-hop multimodal question-answer pairs into atomic multi-hop claims and generate a set of candidate claims. These candidate claims are then modified to include additional hops and refined for fluency and clarity according to a set of annotation guidelines. To ensure the accuracy of the claims, we use a Retrieval-Augmented Generation (RAG)-based validation method to verify their validity. Finally, we ask a group of human annotators to score the claims based on their fluency, correctness, and clearness, and manually rewrite the claims that are below a certain threshold.

We establish performance baselines on \textsc{MMCV} using three state-of-the-art multimodal large language models (MLLMs) and highlight their limitations in verifying complex multimodal claims. We further demonstrate the challenges posed by the dataset, especially as the number of reasoning hops increases, by illustrating the constrained performance of various prompt techniques designed to enhance MLLMs' reasoning capabilities, including chain-of-thought, self-ask, and symbolic-guided reasoning. Additionally, we establish a human performance benchmark on a subset of MMCV.

Overall, we introduce a challenging multi-hop multimodal claim verification dataset that includes claims with up to 4 reasoning hops. These complex claims often consist of multiple sentences linked by coreference and demand evidence from various modalities, such as text, images, and tables. Table \ref{tab:comparison} provides a comparison between \textsc{MMCV} and existing popular claim verification datasets. While current datasets typically focus on either multimodal claims or multi-hop textual claims, none of them incorporate multi-hop multimodal claims that necessitate cross-modal reasoning. We hope that the introduction of \textsc{MMCV} and its corresponding evaluation task will inspire further research in complex multi-hop multimodal reasoning for claim verification. 
In summary, our contributions include:
\begin{itemize}
    \item We introduce and formalize the multi-hop multimodal claim verification task.
    \item We develop a novel pipeline that leverages LLMs for data annotation, enhanced by human feedback, to construct a benchmark dataset for multi-hop multimodal claim verification. This method significantly lowers the cost and labor required to produce a large-scale dataset.
    \item We establish baseline performance on this task using MLLMs and human evaluation. Our analysis shows that this is a non-trivial task, with several challenges that remain to be addressed in future work.
\end{itemize}
\section{Background}
\label{sec:background}

\noindent \textbf{Multimodal Claim Verification.}
Previous research on claim verification has primarily focused on textual data. However, with the growing recognition that misinformation often appears across multiple modalities and that multimodal misinformation is perceived as more credible and spreads faster than text-only misinformation, recent efforts have shifted toward verifying multimodal claims \cite{akhtar-etal-2023-multimodal}. As a result, several multimodal claim verification datasets have been proposed including FakeNewsNet \cite{shu2020fakenewsnet}, COSMOS \cite{aneja2021cosmos}, InfoSurgeon \cite{fung-etal-2021-infosurgeon}, Factify \cite{mishra2022factify}, Fauxtography \cite{zlatkova-etal-2019-fact}, and Mocheg \cite{yao2023end}. However, to the best of our knowledge, there are no existing datasets for multi-hop multimodal claim verification, which challenges the system's reasoning capability by requiring it to integrate and interpret multiple pieces of evidence from different modalities.
\\

\noindent \textbf{Multi-hop Reasoning.}
Verifying complex claims often requires multi-step (multi-hop) reasoning \cite{mavi2022survey}, which requires combining information from multiple pieces of evidence to predict the veracity of a claim. Many recently proposed datasets are created to challenge a model's ability to reason across multiple sentences or documents. These include MultiRC \cite{khashabi-etal-2018-looking}, QAngaroo \cite{welbl2018constructing}, ComplexWebQuestion \cite{talmor2018web}, HotpotQA \cite{yang-etal-2018-hotpotqa}, and HoVer \cite{jiang-etal-2020-hover}. In contrast to these datasets, \textsc{MMCV} incorporates context from various modalities, such as images and tables, further challenging the system's ability to understand and integrate evidence from different sources.
\\

\noindent \textbf{Construct Synthetic Dataset with LLMs.}
The emergence of advanced large language models has sparked growing interest in automating the data annotation process using LLMs \cite{tan2024large, wu2024unigen, bao2024autobench, chen2024gui}, driven by their advanced capabilities, including in-context learning \cite{dong2022survey} and learning from human feedback \cite{ouyang2022training}. \cite{wang-etal-2023-lets} propose an explain-then-generate pipeline using LLMs for iterative data synthesis, while \cite{pace2024west} combine the Best-of-N and Worst-of-N sampling strategies to introduce the West-of-N approach. With this same objective, the multi-hop claims in \textsc{MMCV} are created and refined by LLMs using human feedback, following guidelines and rules specifically designed to enforce a multi-hop structure within each claim.
\\
\begin{figure*}[!tbp]
    \centering
    \includegraphics[width=0.99\linewidth]{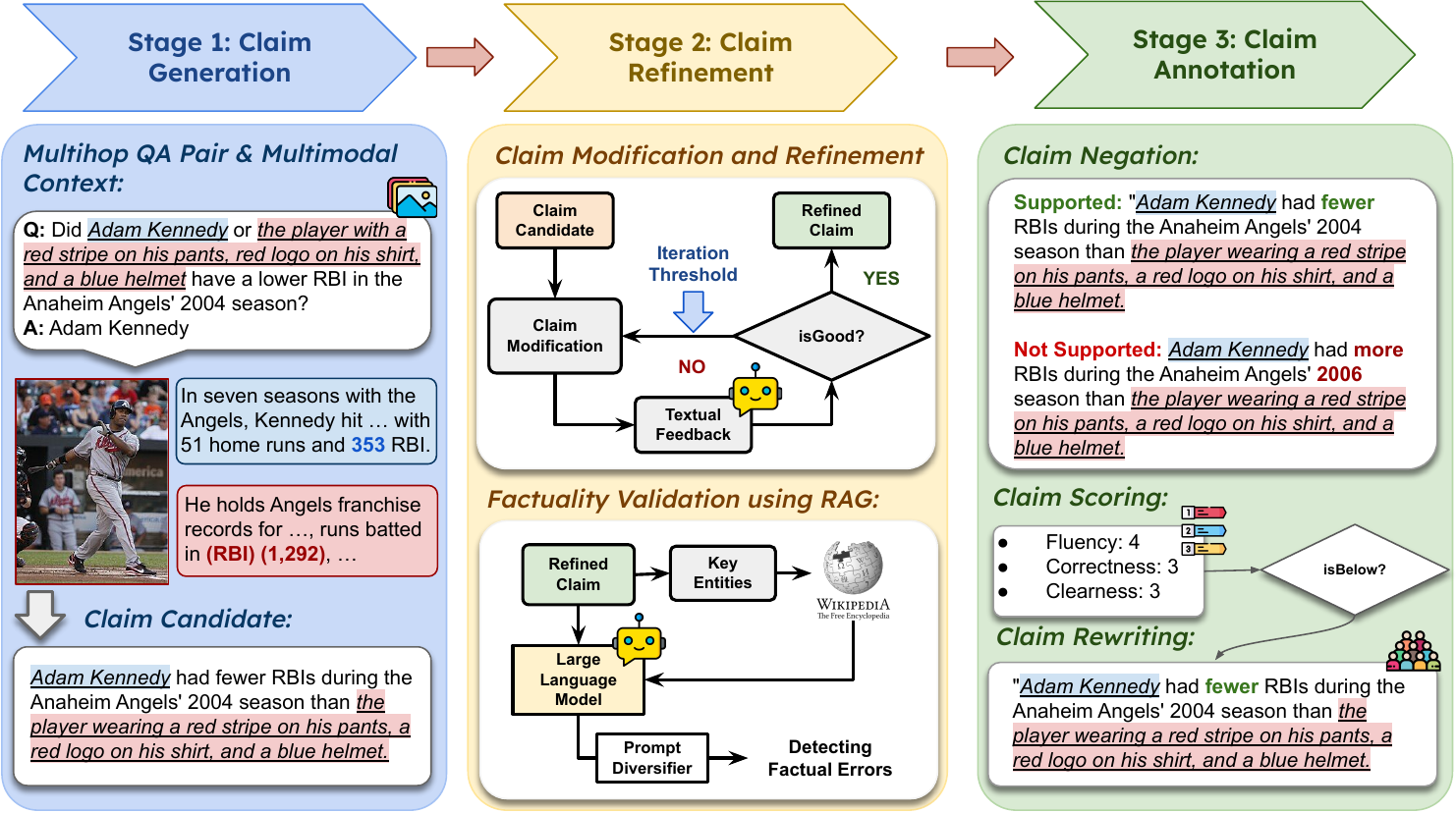}
    \caption{
    Overview of data collection flow chart for \textit{MMCV}. In the first stage, we re-formulate question-answer pairs from \textit{MultimodalQA} to generate candidate claims. In the second stage, we modify and refine the candidate claims, and apply a Retrieval-Augmented Generation (RAG)-based method to verify their correctness. In the final stage, we ask human annotators to rank the candidate claims to select the best one and label the final claims accordingly.
    }
    \label{fig:pipeline}
\end{figure*}

\section{The \textsc{MMCV} dataset}
\label{sec3:dataset}
The main goal of our work is to compile a diverse and extensive collection of multi-hop claims that require joint reasoning across evidence from different modalities, such as text, tables, and images, for verification. One approach to achieving this is to transform multimodal question-answering pairs into atomic claims and refine them to incorporate additional reasoning steps, making them more natural. However, there are two major challenges in creating such a dataset: first, \textit{building a large-scale dataset is labor-intensive and costly}; second, in our pilot studies, we found that simply providing instructions to crowd workers and asking them to rewrite multi-hop claims is counterproductive, as \textit{it is difficult to control quality and challenging for workers to create meaningful multi-hop claims}. Instead, we develop a pipeline that leverages the emerging capabilities of large language models to generate text and learn from feedback, with human input to ensure the quality of the final output. 

In this approach, LLMs handle the mundane task of rewriting claims consistently according to the instructions, while human effort is significantly reduced to quality control of the final claims based on a set of guidelines. Figure \ref{fig:pipeline} shows the overall workflow of our data construction pipeline, which contains three stages: LLM-Based Claim Generation (\S \ref{sec:claim_generation}), LLM-Generated Claim Refinement (\S \ref{sec:claim_refinement}) and Claim Annotation by Human (\S \ref{sec:claim_annotation}).

\subsection{Claim Generation}
\label{sec:claim_generation}
In this stage, we leverage the in-context learning capabilities of large language models to transform question-answer pairs from the MultimodalQA dataset \cite{talmor2021multimodalqa} into verifiable claims. To minimize the impact of in-context examples on the quality of the generated claims, we carefully craft a pool of 20 in-context examples and randomly select 3 for use during execution. The claims are formulated to ensure that no information is omitted from the original QA pairs and no new information is introduced. Since the claims are derived directly from the question and the correct answer, they are automatically labeled as \textsc{SUPPORT}. The prompt template for claim generation is listed in Appendix \ref{sec:experiment_prompt}.

\subsection{Claim Refinement}
\label{sec:claim_refinement}
After generating the initial claims from the question-answer pairs, we modify and refine them to ensure they are more naturally phrased and more accurately supported by the facts. Next, we review the claims for any factual errors that may have been introduced during the modification process and make corrections as needed.

\noindent \textbf{Claim Modification and Refinement.}
To introduce additional reasoning steps to the claim candidate, we employ a modify-then-refine approach that iteratively enhances the quality of the modified claim candidate based on feedback from LLMs \cite{pan2023automatically}. Specifically, we begin by identifying the Wikipedia entities mentioned in the answers from the question-answer pairs. If there is only one Wikipedia entity in the answer, we leave the claim candidate unchanged. However, if there are multiple Wikipedia entities, we use the summaries of their respective Wikipedia articles as context and instruct the LLMs to modify the claim in such a way that it incorporates this contextual information to replace the entity, ensuring that the entity's name does not appear directly in the claim.

To help LLMs understand the modification task, we provide them with 3-5 randomly selected in-context examples from a pool of hand-crafted examples. After modifying the claim, we obtain feedback from LLMs regarding the fluency, correctness, and clarity of the modified claim. The criteria used for this assessment are listed in the Appendix \ref{sec:experiment_prompt}. If the feedback suggests further improvement, the claim is sent back to the modification step, incorporating the LLMs' feedback until a certain iteration threshold is reached. If the modified claim still does not pass the quality check, it is marked for manual review and revision by human annotators.

\noindent \textbf{RAG-based Truthfulness Validation.}
Since we introduce additional contextual information from Wikipedia when modifying the claims, there is a risk that LLMs might hallucinate and produce outputs that are not faithful to the input context. To eliminate potential factual errors, we use a retrieval-augmented generation (RAG) \cite{lewis2020retrieval}-based pipeline to retrieve the full Wikipedia articles of the relevant entities and validate the factual accuracy of the modified claims. To mitigate the impact of prompt sensitivity on the model's output \cite{lu-etal-2022-fantastically, sclar2023quantifying}, we diversify the prompts by randomly changing their format for each verification step. For instance, instead of consistently using \texttt{Is it true that \{claim\}?}, the prompt is randomly chosen from a set of equivalent alternatives, such as \texttt{Verify the following statement: \{claim\}} or \texttt{What evidence supports the claim that \{claim\}?}

\subsection{Claim Annotation}
\label{sec:claim_annotation}
At this stage, we have obtained claims that have been modified and refined by LLMs and factually validated by RAG-based pipelines. Next, we use LLMs to generate negated claims by applying a set of specific negation rules. We employ three distinct methods for generating these negated claims. For instance, given the claim,  \textit{``Since its construction in 1889, the Eiffel Tower in Paris attracts millions of visitors annually.''}, the results after applying the negation rules are as follows:

\begin{tcolorbox}[negation]
\small{
    $\triangleright$ \textbf{Word substitution:} \textit{The Eiffel Tower in Paris houses millions of residents annually.} \\ 
    $\triangleright$ \textbf{Entity substitution:} \textit{The Colosseum in Paris attracts millions of visitors annually.} \\ 
    $\triangleright$ \textbf{Temporal mutation:} \textit{Ever since its construction in 2050, the Eiffel Tower has been Paris's top tourist site.} 
}
\end{tcolorbox}

Next, a group of human annotators is tasked with evaluating the claims based on three dimensions: fluency, correctness, and clarity, scoring each dimension on a scale of 1 to 5. Fluency assesses how naturally the claim reads, as outputs generated by language models can sometimes sound artificial. Correctness evaluates whether the claim is factually accurate based on the evidence. Clarity determines if the claim is easily understood, as entity substitution might make it difficult to comprehend. Once the claims are scored, the average of the fluency, correctness, and clarity scores is calculated to determine the final score for each claim. If a claim's final score falls below a predetermined threshold, it is flagged and sent back to the annotators for manual revision. Detailed annotation guidelines are listed in Appendix \ref{sec:annotation_guidelines}.

\section{Dataset Analysis}
\label{sec:analysis}

\noindent \textbf{Dataset Statistics.}
\textsc{MMCV} contains 15,569 multi-hop multimodal claims, with their statistics detailed in Table \ref{tab:data_stat}. The number of hops is determined by the count of multimodal evidence associated with each claim. The dataset includes a balanced distribution of SUPPORT and REFUTE claims. Specifically, there are 5,884 1-hop claims with an average of 21.7 tokens per claim; 8,485 2-hop claims averaging 25.32 tokens per claim; 804 3-hop claims with an average of 25.44 tokens per claim; and 396 4-hop claims averaging 26.17 tokens per claim. An example from the dataset is provided in Appendix \ref{sec:dataset_example}.\\
\\
\noindent \textbf{Multi-hop Reasoning Types.}
We provide examples of each reasoning type in Table \ref{tab:dataset_example}. Most 1-hop and 2-hop claims require at least one supporting fact from either image or table evidence for verification. In contrast, the majority of 3-hop and 4-hop claims require evidence from all three modalities. The process of removing a bridge entity and replacing it with a relative clause or phrase significantly increases the informational load of a single hypothesis. As a result, some 3-hop and 4-hop claims are relatively longer and exhibit complex syntactic and reasoning structures. Our experimental results also indicate that the difficulty for models to verify claims escalates as the hop count increases.

\begin{table}[t]
\renewcommand{\arraystretch}{1.3}%
\centering
\resizebox{\linewidth}{!}{%
\begin{tabular}{@{}lllll@{}}
\toprule
\textbf{Data }                                & \textbf{1-hop} & \textbf{2-hop} & \textbf{3-hop} & \textbf{4-hop} \\ \hline
\# Claims                            & 5,884      & 8,485      & 804      & 396 \\
Ave. \# Tokens in Claim              & 21.7       & 25.32      & 25.44    & 26.17 \\
Max. \# Tokens in Claim              & 48         & 58         & 51       & 63 \\ \hline
\# Text Evidence                     & 2,590      & 7,323      & 1,142    & 760 \\
\# Image Evidence                    & 1,979      & 2,948      & 634      & 512 \\
\# Table Evidence                    & 1,315      & 6,699      & 636      & 312 \\ \hline
\# \textsc{SUPPORT} Labels           & 2,824      & 4,030      & 349      & 158 \\
\# \textsc{REFUTE} Labels            & 3,060      & 4,455      & 455      & 238 \\ \bottomrule
\end{tabular}
}
\caption{Dataset Statistics of \textsc{MMCV}.}
\label{tab:data_stat}
\end{table}

\begin{table*}[t]
\renewcommand{\arraystretch}{1.2}%
\centering
\resizebox{\linewidth}{!}{%
\begin{tabular}{@{}cccccccccccccc@{}}
\toprule
                             &        & \multicolumn{3}{c}{\textbf{1-hop}} & \multicolumn{3}{c}{\textbf{2-hop}} & \multicolumn{3}{c}{\textbf{3-hop}} & \multicolumn{3}{c}{\textbf{4-hop}} \\ \midrule
\textbf{Retrieval} & \textbf{Model}  & P     & R     & F1    & P     & R     & F1    & P     & R     & F1    & P     & R     & F1    \\ \midrule
\multirow{3}{*}{\textit{Closed-book}} & \textsc{GPT-4o} & 76.86   & 72.94  & \color{OliveGreen}{\textbf{\underline{71.79}}}  & 67.96   & 63.30  & 60.66  & 62.88   & 58.89  & 56.17  & 67.93   & 62.39  & 61.20  \\
          & \textsc{Gemini} & 75.67 & 71.44 & 70.15 & 69.10 & 64.19 & 61.73 & 66.74 & 61.10 & 58.44 & 63.78 & 59.90 & 58.69 \\
          & \textsc{LLaVA}  & 64.18 & 63.78 & 63.57 & 64.06 & 63.93 & \color{OliveGreen}{\textbf{\underline{63.87}}} & 66.78 & 66.81 & \color{OliveGreen}{\textbf{\underline{66.76}}} & 64.64 & 64.84 & \color{OliveGreen}{\textbf{\underline{64.64}}} \\ \midrule
\multirow{3}{*}{\textit{Open-book}}   & \textsc{GPT-4o} & 76.95   & 72.95  & 71.78  & 68.03   & 63.24  & 60.53  & 62.67   & 58.78  & 56.08  & 67.75   & 62.46  & 61.35  \\
          & \textsc{Gemini} & 79.58 & 79.25 & \color{OliveGreen}{\textbf{\underline{79.20}}} & 72.38 & 71.85 & \color{OliveGreen}{\textbf{\underline{71.66}}} & 66.37 & 65.90 & \color{OliveGreen}{\textbf{\underline{65.86}}} & 67.21 & 66.86 & \color{OliveGreen}{\textbf{\underline{66.97}}} \\
          & \textsc{LLaVA}  & 62.86 & 59.68 & 57.21 & 64.17 & 62.48 & 61.50 & 65.47 & 64.64 & 63.76 & 66.50 & 66.76 & 66.42 \\ \bottomrule
\end{tabular}
}
\caption{We report the Precision, Recall, and F1 scores of various MLLMs on MMCV for zero-shot multimodal claim verification. In the closed-book setting, the model verifies the claim without access to any external knowledge sources. In the open-book setting, the model is provided with a set of gold evidence. The best-performing model for each hop is highlighted in {\color{OliveGreen}{\textbf{\underline{Green}}}} for both settings.}
\label{tab:mllm_exp}
\end{table*}

\section{Experiments and Results}
\label{sec:experiments}
In this section, we discuss our experiment settings (\S \ref{sec:experiment_settings}), the experiment results (\S \ref{sec:results}), and the error analysis (\S \ref{sec:error_analysis}). We begin by formally defining the MMCV task below.\\
\\
\noindent \textbf{Task Definition.}
The formulation of multi-hop multimodal claim verification is defined as follows: Given a claim $C$, and a list of multimodal evidence $\mathcal{E}(C)$, which includes text, images, and tables, the system must reason over all the evidence and predict the label of the claim as either \textsc{SUPPORT} or \textsc{REFUTE}. 

\subsection{Experiment Settings}
\label{sec:experiment_settings}
As there are no existing models specifically designed for multi-hop multimodal supervised claim verification, we conduct our experiments using MLLMs. Moreover, previous studies in textual claim verification and multimodal claim verification indicate that LLMs and MLLMs can significantly enhance task performance compared to traditional supervised approaches \cite{pan-etal-2023-fact, wang-shu-2023-explainable, li-etal-2024-self, geng2024multimodal}. Furthermore, supervised methods often require extensive annotated corpora, which are difficult to acquire and limit domain transferability, as training data typically covers only a single domain.

\noindent \textbf{Zero-shot Claim Verification.}
We establish performance baselines for zero-shot multimodal claim verification using various MLLMs under two settings. In the \textit{closed-book setting}, the model does not retrieve information from external knowledge sources and must rely on its parametric (internal) knowledge to verify the claim. In the \textit{open-book setting}, the model is provided with a set of gold evidence. Specifically, we use the prompt from \cite{geng2024multimodal}, which extracts the models' predictions, explanations, and confidence levels. The prompt is listed in Appendix \ref{sec:experiment_prompt}. We use macro precision, recall, and F-1 score to evaluate the model performance. \\
\\
\noindent \textbf{MLLM.}
We utilize two state-of-the-art MLLMs: GPT-4o \cite{achiam2023gpt} and Gemini 1.5 Flash \cite{team2023gemini}. Additionally, we evaluate the performance of an open-source MLLM, LLaVA-V1.5-7B \cite{liu2024visual}, on MMCV. The temperature is set to 0.0, and the maximum number of tokens is set to 5000. \\
\\
\noindent \textbf{Prompts for Enhanced Reasoning} In addition to the prompt mentioned above, we conduct experiments using specialized prompting techniques aimed at eliciting reasoning from LLMs, such as Chain-of-Thought \cite{wei2022chain} and Self-Ask \cite{press-etal-2023-measuring}. We also test symbolic-guided reasoning prompts like ProgramFC \cite{pan-etal-2023-fact} and Visual Programming \cite{gupta2023visual}. To minimize the overall cost of the experiments, we randomly select 100 examples from each hop of the MMCV dataset for testing. The experiments are conducted using open-book setting.\\
\\
\noindent \textbf{Human Performance}
To benchmark human performance on our dataset, we used the same randomly selected examples employed in the enhanced reasoning prompt experiments. We recruited four experts in automated fact-checking research to classify multihop claims from MMCV based on the provided evidence. The SMART \cite{chew2019smart} framework \footnote{\href{https://github.com/RTIInternational/SMART}{https://github.com/RTIInternational/SMART}} was used to deploy the annotation task, and human performance was evaluated using the macro F-1 score.

\begin{figure*}[!t]
  \centering
  \resizebox{0.99\linewidth}{!}{%
  \begin{subfigure}[b]{0.5\textwidth}
    \centering
    \includegraphics[width=\textwidth]{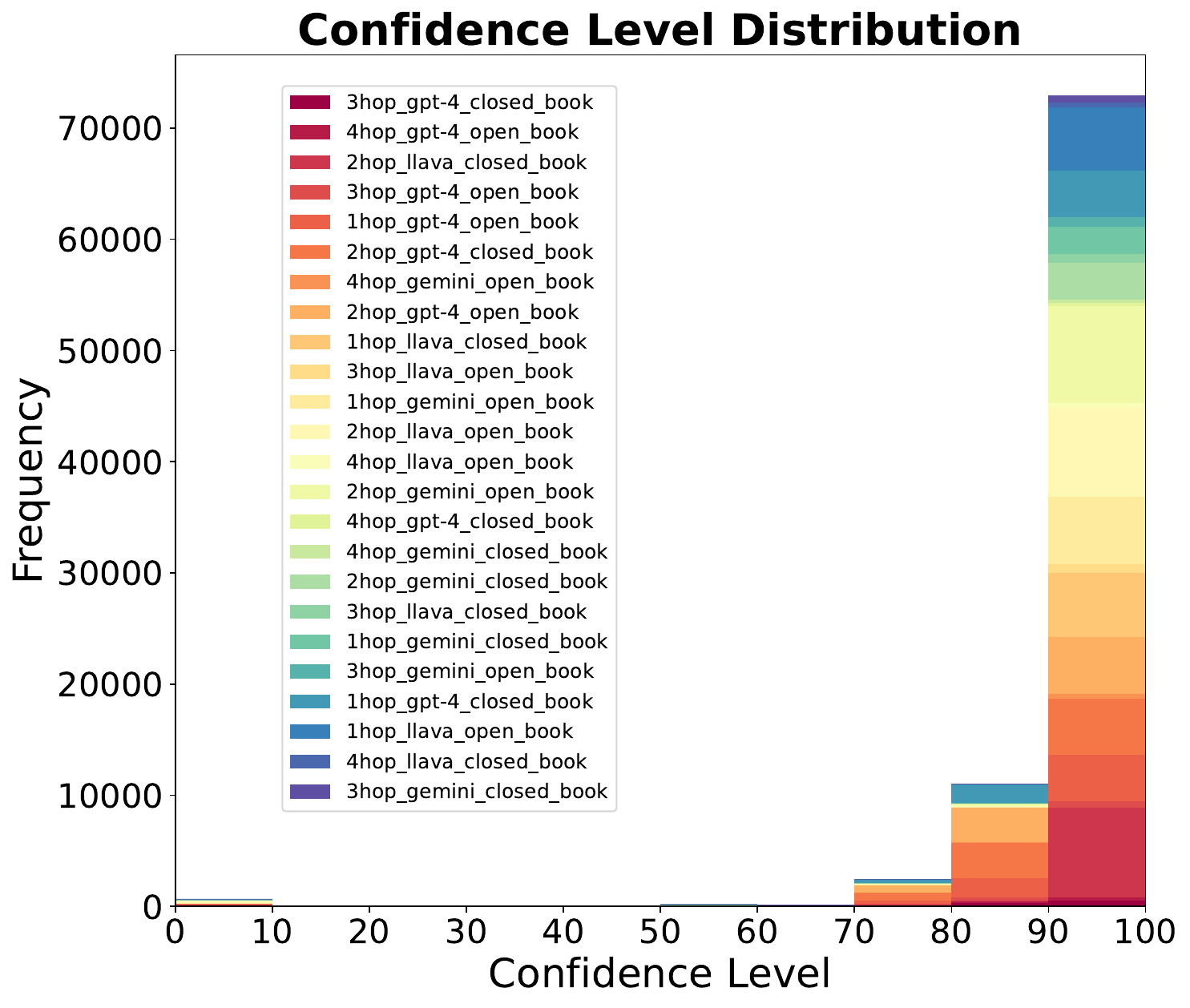}
    \label{fig:pol_compare}
  \end{subfigure}
  \begin{subfigure}[b]{0.5\textwidth}
    \centering
    \includegraphics[width=\textwidth]{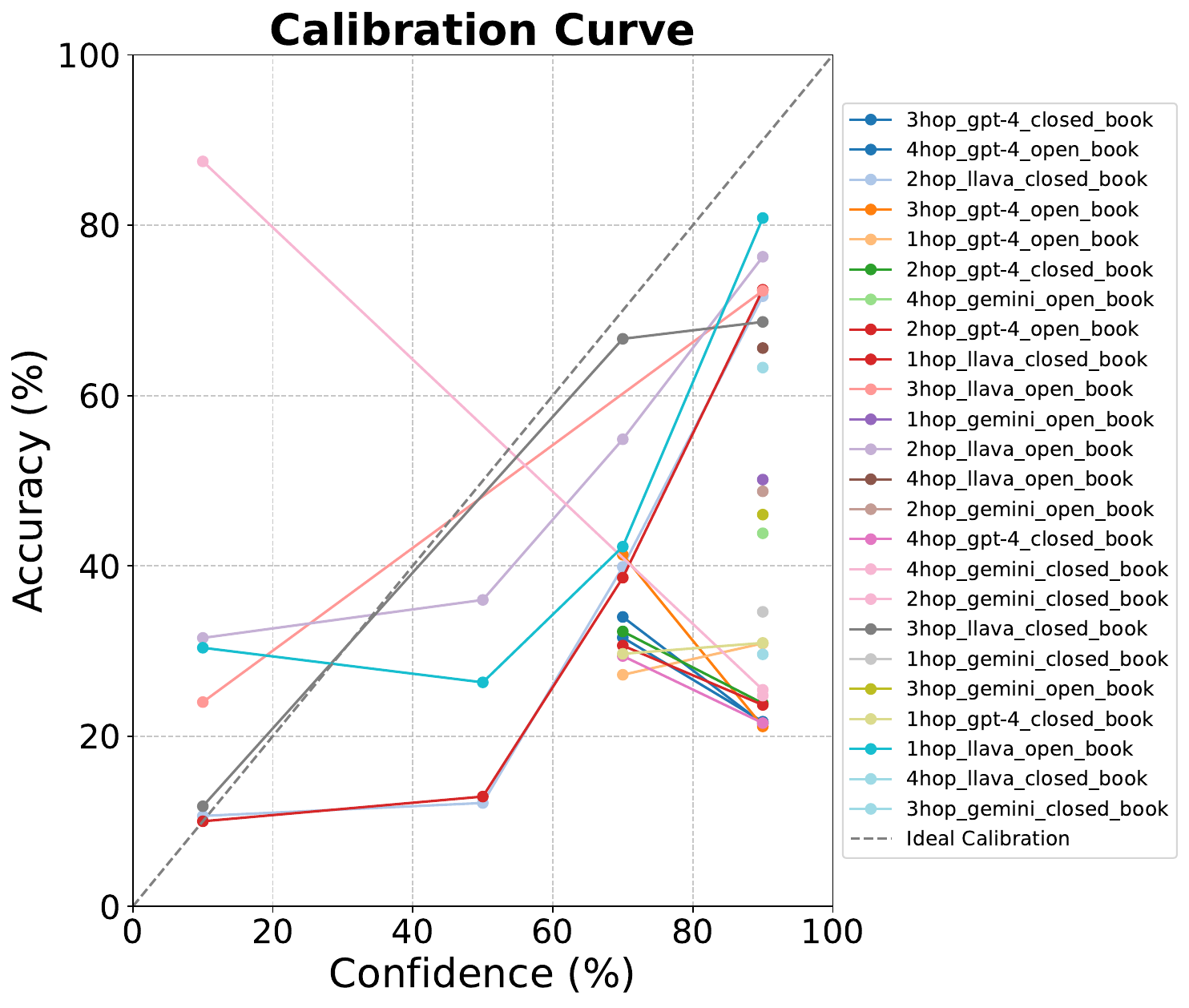}
    \label{fig:gos_compare}
  \end{subfigure}}
  \vspace{-0.2cm}
  \caption{The left figure shows the confidence score distribution of GPT4-o, Gemini, and LLaVA on MMCV under both open-book and closed-book settings, categorized by the number of hops. The right figure shows their calibration curves.}
  \label{fig:confidence}
\end{figure*}

\subsection{Experiment Results}
\label{sec:results}
\noindent \textbf{Main Results.}
We report the comprehensive results of the three MLLMs on MMCV in Table \ref{tab:mllm_exp}, highlighting the best-performing models for each hop under both open-book and closed-book settings. Overall, Gemini 1.5 outperforms others in the open-book setting with an average F-1 of 70.92, while LLaVA achieves the highest performance in the closed-book setting with an average F-1 of 66.77. This is surprising, given that LLaVA is a much smaller model compared to GPT4-o and Gemini, and therefore possesses less parametric knowledge. Upon manually analyzing a subset of 100 randomly selected outputs from LLaVA, we found that the model frequently hallucinates, even when it predicts the correct label, particularly as the hop count increases. This is consistent with its open-book performance, where its accuracy declines when provided with gold evidence. Additionally, we observe that GPT4-o performs slightly better in closed-book settings than in open-book settings, suggesting a tendency to hallucinate. In contrast, Gemini's performance drops significantly in closed-book settings compared to open-book, demonstrating its robustness in effectively utilizing provided gold evidence.\\
\\
\noindent \textbf{Confidence Level Analysis}
The left panel of Figure \ref{fig:confidence} presents the confidence \cite{geng2024survey} distributions for all three MLLMs, categorized by the number of hops and divided into 10 intervals. The results show that the majority of the MLLMs are concentrated in the 90-100 confidence range, with only a small number exhibiting low confidence (0-10 range), which occurs solely in open-book settings. This indicates that the MLLMs consider the provided gold evidence.

The right panel of Figure \ref{fig:confidence} displays the calibration curves, illustrating the relationship between the models' confidence levels and their actual classification accuracy. These curves reveal a positive correlation between confidence and accuracy for 1-hop and 2-hop claims, as exemplified by the red line (GPT-4-o on 2-hop), the teal line (LLaVA on 1-hop), and the purple line (Gemini on 1-hop). In contrast, the downward curves, mostly observed in 3-hop and 4-hop claims, suggest that the models tend to be overconfident when classifying more complex claims. Additionally, the results indicate that open-book settings generally have better-calibrated confidence scores than closed-book settings, further suggesting that the models exhibit overconfidence when not provided with gold evidence.\\

\begin{table}[!t]
\renewcommand\arraystretch{1.3}
\resizebox{0.99\linewidth}{!}{%
\begin{tabular}{cccccc}
\toprule
\textbf{Model} & \textbf{Method} & \textbf{1-hop} & \textbf{2-hop} & \textbf{3-hop} & \textbf{4-hop} \\ \midrule
\multirow{3}{*}{\textsc{Gemini 1.5}}    & CoT           & 78.52  & 69.66  & 67.45  & 70.24  \\
                        & Self-Ask      & 75.47  & 66.58  & 60.94  & 70.67  \\
                        & Symbolic & 74.89  & 63.82  & 54.61  & 72.36  \\ \midrule
\multirow{3}{*}{\textsc{GPT4-o}} & CoT          & 80.43  & 83.33  & 71.20  & 72.99  \\
                        & Self-Ask      & 77.42  & 80.12  & 70.52  & 75.23  \\
                        & Symbolic & 80.56  & 78.78  & 68.72  & 75.67  \\ \bottomrule
\end{tabular}
}
\caption{Results of Gemini and GPT4-o on 100 randomly sampled claims for each hop using three types of reasoning prompts. Model performance is evaluated using F-1 score.}
\label{tab:reasoning_prompt}
\end{table}

\noindent \textbf{Reasoning Prompt Results.}
Table \ref{tab:reasoning_prompt} reports the performance of Gemini and GPT4-o on the randomly sampled subset of MMCV under open-book settings using various prompts that elicit LLMs' reasoning abilities. For symbolic approach, we ask LLMs to first generate a Python-like program that decomposes the mutli-hop claim into a set of function calls that describe the reasoning steps required to verify the claim, and use the symbolic information provided by the generated program to elicit better step-by-step reasoning from the model. We observe that GPT-4-o gains more from the enhanced reasoning prompt compared to Gemini, achieving a higher average F1 score of 75.93 in symbolic guided reasoning, whereas Gemini attains an average F1 score of 66.42 for the same task. Additionaly, we found that Symbolic approach are more effective on 4-hop claims, having a higher F1 score than CoT and self-ask. However, this observation is different on simpler 2-hop and 3-hop claims, where CoT appears to be more effective.\\

\begin{table}[!t]
\centering
\resizebox{0.99\linewidth}{!}{%
\begin{tabular}{ccccc}
\toprule
\textbf{Annotator} & \multicolumn{4}{c}{\textbf{\# Hops}}    \\ 
\midrule
    & 1-hop & 2-hop & 3-hop & 4-hop \\ 
\midrule
\textit{Annotator 1} & 83.33 & 86.20 & 78.42 & 79.82  \\
\textit{Annotator 2} & 82.46 & 88.29 & 79.45 & 82.16  \\
\textit{Annotator 3} & 80.60 & 90.53 & 80.62 & 85.24  \\
\textit{Annotator 4} & 79.64 & 86.50 & 82.32 & 83.87  \\
\bottomrule
\end{tabular}
}
\caption{Results of human performance on 200 random samples. Performance are measured by F-1 score.}
\label{tab:human_performance}
\end{table}

\noindent \textbf{Human Performance Results}
\label{sec:human_performance}
To establish human performance on our dataset, we randomly sampled 200 examples, with 50 examples from each hop from MMCV. We recruited four annotators to perform claim verification given the gold evidence. We trained our annotators on the task by providing them with guidelines and sample annotations to ensure consistency and accuracy in their evaluations. After training, the annotators independently verified each claim using the provided gold evidence, allowing us to assess the human baseline performance on the dataset.
Table \ref{tab:human_performance} reports the results from the human annotators. We observe that the human annotators achieve very high performance in verifying the claims across all 4 hops. The human performance is 23.3\% and 27.3\% higher than the best-performing MLLMs on 3-hop and 4-hop claims respectively. This suggests that although MLLMs perform relatively well, there is still room for improvement to match human performance.

\subsection{Error Analysis}
\label{sec:error_analysis}
Figure \ref{fig:gpt4_error}, \ref{fig:gemini_error}, and \ref{fig:llava_error} shows the error analysis of the false positive examples from GPT4-o, Gemini, and LLaVA respectively. We observe that visual misinterpretation is a major issue, with the system often misidentifying or miscontextualizing image elements. This problem is especially pronounced in examples involving sports logos and movie posters, highlighting the need for improvements in the visual processing component.

Another notable issue is the system's handling of temporal and factual information. Errors related to player career timelines and historical events reveal shortcomings in temporal reasoning and the integration of world knowledge. The system's confidence levels, often between 80\% and 100\% for incorrect predictions, suggest a miscalibration in certainty estimation. This overconfidence in erroneous conclusions highlights the need for a more refined approach to confidence scoring.

Last but not least, examples from higher hop categories reveal significant weaknesses in handling complex reasoning tasks. The system often struggles with multi-step logical inferences, frequently failing to coherently link disparate pieces of information. This limitation is especially problematic for claims that require advanced analysis or the cross-referencing of multiple facts.

\section{Conclusion}
In this paper, we introduce \textsc{MMCV}, a multi-hop multimodal claim verification dataset that requires models to aggregate information from up to four multimodal evidence to verify a claim. To create this large-scale dataset, we developed a novel data collection pipeline that leverages the capabilities of LLMs combined with human feedback. Specifically, our approach includes a module that iteratively refines modified claims using feedback from a judge LLM based on a set of predefined criteria, as well as an actuality validation module that employs RAG to ensure the factual accuracy of the claims.
Our results show that state-of-the-art MLLMs struggle to verify more complex claims as the number of reasoning hops increases, often displaying overconfidence in their predictions. We also present findings from experiments utilizing prompts tailored to enhance the reasoning abilities of MLLMs, alongside human performance benchmarks for comparison. Additionally, we categorize and provide a detailed error analysis of false positive results from each model.
We hope that \textsc{MMCV} will inspire the development of models capable of conducting complex, multi-hop reasoning in the challenging task of multimodal claim verification.

\section{Limitations}
We identify two main limitations of MMCV.
First, the construction of MMCV depends on in-context learning coupled with self-refinement to convert a natural language question-answer pair into a multi-hop claim. While this method has proven to be effective, it may face difficulties when dealing with questions with intricate grammar structures and logical structures. This arises from the difficulty in conveying complex grammatical rules to the language model through a limited number of demonstrations within a constrained context size.
Second, our aggregation method purely relies on LLMs themselves, which could introduce potential hallucination problems. On the other hand, by using a more robust logic solver could help with the hallucination issues, but there would be a tradeoff between the applicability and the robustness of the model.

\section{Ethical Statement}
\noindent \textbf{Biases.} We acknowledge the possibility of biases existing within the data used for training the language models, as well as in certain factuality assessments. Unfortunately, these factors are beyond our control.

\noindent \textbf{Intended Use and Misuse Potential.} Our models have the potential to verify complex multimodal claims. However, it is essential to recognize that they may also be susceptible to misuse by malicious individuals. Therefore, we strongly urge researchers to approach their utilization with caution and prudence.

\noindent \textbf{Environmental Impact.} We want to highlight the environmental impact of using large language models, which demand substantial computational costs and rely on GPUs/TPUs for training, which contributes to global warming. 
However, it is worth noting that our approach does not train such models from scratch. Instead, we use few-shot in-context learning.
Nevertheless, the large language models we used in this paper are likely running on GPU(s).

\section*{Acknowledgements}
This material is based upon work supported by the U.S. Department of Homeland Security under Grant Award Number 17STQAC00001-07-04, NSF awards (SaTC-2241068, IIS-2339198), a Cisco Research Award, and a Microsoft Accelerate Foundation Models Research Award. The views and conclusions contained in this document are those of the authors and should not be interpreted as necessarily representing the official policies, either expressed or implied, of the U.S. Department of Homeland Security or the National Science Foundation.
This research was partially supported by the National Institute Of Diabetes And Digestive And Kidney Diseases of the National Institutes of Health under Award Number K25DK135913.

\bibliography{anthology,custom}

\begin{thebibliography}{58}
\providecommand{\natexlab}[1]{#1}

\bibitem[{Achiam et~al.(2023)Achiam, Adler, Agarwal, Ahmad, Akkaya, Aleman,
  Almeida, Altenschmidt, Altman, Anadkat et~al.}]{achiam2023gpt}
Josh Achiam, Steven Adler, Sandhini Agarwal, Lama Ahmad, Ilge Akkaya,
  Florencia~Leoni Aleman, Diogo Almeida, Janko Altenschmidt, Sam Altman,
  Shyamal Anadkat, et~al. 2023.
\newblock Gpt-4 technical report.
\newblock \emph{arXiv preprint arXiv:2303.08774}.

\bibitem[{Akhtar et~al.(2023)Akhtar, Schlichtkrull, Guo, Cocarascu, Simperl,
  and Vlachos}]{akhtar-etal-2023-multimodal}
Mubashara Akhtar, Michael Schlichtkrull, Zhijiang Guo, Oana Cocarascu, Elena
  Simperl, and Andreas Vlachos. 2023.
\newblock \href {https://doi.org/10.18653/v1/2023.findings-emnlp.361}
  {Multimodal automated fact-checking: A survey}.
\newblock In \emph{Findings of the Association for Computational Linguistics:
  EMNLP 2023}, pages 5430--5448, Singapore. Association for Computational
  Linguistics.

\bibitem[{Aneja et~al.(2021)Aneja, Bregler, and Nie{\ss}ner}]{aneja2021cosmos}
Shivangi Aneja, Chris Bregler, and Matthias Nie{\ss}ner. 2021.
\newblock Cosmos: Catching out-of-context misinformation with self-supervised
  learning.
\newblock \emph{arXiv preprint arXiv:2101.06278}.

\bibitem[{Bao et~al.(2024)Bao, Huang, Wang, Ye, Wang, Chen, Elhoseiny, and
  Zhang}]{bao2024autobench}
Han Bao, Yue Huang, Yanbo Wang, Jiayi Ye, Xiangqi Wang, Xiuyin Chen, Mohamed
  Elhoseiny, and Xiangliang Zhang. 2024.
\newblock Autobench-v: Can large vision-language models benchmark themselves?
\newblock \emph{arXiv preprint arXiv:2410.21259}.

\bibitem[{Chen et~al.(2024)Chen, Huang, Wu, Tang, Chen, Bai, He, Wang, Zhou, Li
  et~al.}]{chen2024gui}
Dongping Chen, Yue Huang, Siyuan Wu, Jingyu Tang, Liuyi Chen, Yilin Bai,
  Zhigang He, Chenlong Wang, Huichi Zhou, Yiqiang Li, et~al. 2024.
\newblock Gui-world: A dataset for gui-oriented multimodal llm-based agents.
\newblock \emph{arXiv preprint arXiv:2406.10819}.

\bibitem[{Chew et~al.(2019)Chew, Wenger, Kery, Nance, Richards, Hadley, and
  Baumgartner}]{chew2019smart}
Rob Chew, Michael Wenger, Caroline Kery, Jason Nance, Keith Richards, Emily
  Hadley, and Peter Baumgartner. 2019.
\newblock Smart: an open source data labeling platform for supervised learning.
\newblock \emph{Journal of Machine Learning Research}, 20(82):1--5.

\bibitem[{Dong et~al.(2022)Dong, Li, Dai, Zheng, Wu, Chang, Sun, Xu, and
  Sui}]{dong2022survey}
Qingxiu Dong, Lei Li, Damai Dai, Ce~Zheng, Zhiyong Wu, Baobao Chang, Xu~Sun,
  Jingjing Xu, and Zhifang Sui. 2022.
\newblock A survey on in-context learning.
\newblock \emph{arXiv preprint arXiv:2301.00234}.

\bibitem[{Fung et~al.(2021)Fung, Thomas, Gangi~Reddy, Polisetty, Ji, Chang,
  McKeown, Bansal, and Sil}]{fung-etal-2021-infosurgeon}
Yi~Fung, Christopher Thomas, Revanth Gangi~Reddy, Sandeep Polisetty, Heng Ji,
  Shih-Fu Chang, Kathleen McKeown, Mohit Bansal, and Avi Sil. 2021.
\newblock \href {https://doi.org/10.18653/v1/2021.acl-long.133}
  {{I}nfo{S}urgeon: Cross-media fine-grained information consistency checking
  for fake news detection}.
\newblock In \emph{Proceedings of the 59th Annual Meeting of the Association
  for Computational Linguistics and the 11th International Joint Conference on
  Natural Language Processing (Volume 1: Long Papers)}, pages 1683--1698,
  Online. Association for Computational Linguistics.

\bibitem[{Gao et~al.(2024)Gao, Zhang, Chen, Huang, Wu, Fu, Wan, Zhang, and
  Sun}]{gao2024best}
Chujie Gao, Qihui Zhang, Dongping Chen, Yue Huang, Siyuan Wu, Zhengyan Fu, Yao
  Wan, Xiangliang Zhang, and Lichao Sun. 2024.
\newblock The best of both worlds: Toward an honest and helpful large language
  model.
\newblock \emph{arXiv preprint arXiv:2406.00380}.

\bibitem[{Geng et~al.(2024{\natexlab{a}})Geng, Cai, Wang, Koeppl, Nakov, and
  Gurevych}]{geng2024survey}
Jiahui Geng, Fengyu Cai, Yuxia Wang, Heinz Koeppl, Preslav Nakov, and Iryna
  Gurevych. 2024{\natexlab{a}}.
\newblock A survey of confidence estimation and calibration in large language
  models.
\newblock In \emph{Proceedings of the 2024 Conference of the North American
  Chapter of the Association for Computational Linguistics: Human Language
  Technologies (Volume 1: Long Papers)}, pages 6577--6595.

\bibitem[{Geng et~al.(2024{\natexlab{b}})Geng, Kementchedjhieva, Nakov, and
  Gurevych}]{geng2024multimodal}
Jiahui Geng, Yova Kementchedjhieva, Preslav Nakov, and Iryna Gurevych.
  2024{\natexlab{b}}.
\newblock Multimodal large language models to support real-world fact-checking.
\newblock \emph{arXiv preprint arXiv:2403.03627}.

\bibitem[{Guo et~al.(2022)Guo, Schlichtkrull, and
  Vlachos}]{guo-etal-2022-survey}
Zhijiang Guo, Michael Schlichtkrull, and Andreas Vlachos. 2022.
\newblock \href {https://doi.org/10.1162/tacl_a_00454} {A survey on automated
  fact-checking}.
\newblock \emph{Transactions of the Association for Computational Linguistics},
  10:178--206.

\bibitem[{Gupta and Kembhavi(2023)}]{gupta2023visual}
Tanmay Gupta and Aniruddha Kembhavi. 2023.
\newblock Visual programming: Compositional visual reasoning without training.
\newblock In \emph{Proceedings of the IEEE/CVF Conference on Computer Vision
  and Pattern Recognition}, pages 14953--14962.

\bibitem[{Huang et~al.(2024{\natexlab{a}})Huang, Chen, and
  Shu}]{huang2024authorship}
Baixiang Huang, Canyu Chen, and Kai Shu. 2024{\natexlab{a}}.
\newblock Authorship attribution in the era of llms: Problems, methodologies,
  and challenges.
\newblock \emph{arXiv preprint arXiv:2408.08946}.

\bibitem[{Huang et~al.(2024{\natexlab{b}})Huang, Chen, and Shu}]{huang2024can}
Baixiang Huang, Canyu Chen, and Kai Shu. 2024{\natexlab{b}}.
\newblock Can large language models identify authorship?
\newblock \emph{arXiv preprint arXiv:2403.08213}.

\bibitem[{Huang et~al.(2024{\natexlab{c}})Huang, Yuan, Zhou, Guo, Wang, Zhuang,
  Sun, Sun, Wang, Ye et~al.}]{huang2024social}
Yue Huang, Zhengqing Yuan, Yujun Zhou, Kehan Guo, Xiangqi Wang, Haomin Zhuang,
  Weixiang Sun, Lichao Sun, Jindong Wang, Yanfang Ye, et~al.
  2024{\natexlab{c}}.
\newblock Social science meets llms: How reliable are large language models in
  social simulations?
\newblock \emph{arXiv preprint arXiv:2410.23426}.

\bibitem[{Jiang et~al.(2020)Jiang, Bordia, Zhong, Dognin, Singh, and
  Bansal}]{jiang-etal-2020-hover}
Yichen Jiang, Shikha Bordia, Zheng Zhong, Charles Dognin, Maneesh Singh, and
  Mohit Bansal. 2020.
\newblock \href {https://doi.org/10.18653/v1/2020.findings-emnlp.309}
  {{H}o{V}er: A dataset for many-hop fact extraction and claim verification}.
\newblock In \emph{Findings of the Association for Computational Linguistics:
  EMNLP 2020}, pages 3441--3460, Online. Association for Computational
  Linguistics.

\bibitem[{Jin et~al.(2024{\natexlab{a}})Jin, Choi, Verma, Wang, and
  Kumar}]{jin2024mm}
Yiqiao Jin, Minje Choi, Gaurav Verma, Jindong Wang, and Srijan Kumar.
  2024{\natexlab{a}}.
\newblock Mm-soc: Benchmarking multimodal large language models in social media
  platforms.
\newblock \emph{arXiv preprint arXiv:2402.14154}.

\bibitem[{Jin et~al.(2023)Jin, Lee, Sharma, Ye, Sikka, Divakaran, and
  Kumar}]{jin2023predicting}
Yiqiao Jin, Yeon-Chang Lee, Kartik Sharma, Meng Ye, Karan Sikka, Ajay
  Divakaran, and Srijan Kumar. 2023.
\newblock Predicting information pathways across online communities.
\newblock In \emph{Proceedings of the 29th ACM SIGKDD Conference on Knowledge
  Discovery and Data Mining}, pages 1044--1056.

\bibitem[{Jin et~al.(2022)Jin, Wang, Yang, Sun, Wang, Liao, and
  Xie}]{jin2022towards}
Yiqiao Jin, Xiting Wang, Ruichao Yang, Yizhou Sun, Wei Wang, Hao Liao, and Xing
  Xie. 2022.
\newblock Towards fine-grained reasoning for fake news detection.
\newblock In \emph{Proceedings of the AAAI Conference on Artificial
  Intelligence}, volume~36, pages 5746--5754.

\bibitem[{Jin et~al.(2024{\natexlab{b}})Jin, Zhao, Wang, Chen, Zhu, Xiao, and
  Wang}]{jin2024agentreview}
Yiqiao Jin, Qinlin Zhao, Yiyang Wang, Hao Chen, Kaijie Zhu, Yijia Xiao, and
  Jindong Wang. 2024{\natexlab{b}}.
\newblock Agentreview: Exploring peer review dynamics with llm agents.
\newblock \emph{arXiv preprint arXiv:2406.12708}.

\bibitem[{Khashabi et~al.(2018)Khashabi, Chaturvedi, Roth, Upadhyay, and
  Roth}]{khashabi-etal-2018-looking}
Daniel Khashabi, Snigdha Chaturvedi, Michael Roth, Shyam Upadhyay, and Dan
  Roth. 2018.
\newblock \href {https://doi.org/10.18653/v1/N18-1023} {Looking beyond the
  surface: A challenge set for reading comprehension over multiple sentences}.
\newblock In \emph{Proceedings of the 2018 Conference of the North {A}merican
  Chapter of the Association for Computational Linguistics: Human Language
  Technologies, Volume 1 (Long Papers)}, pages 252--262, New Orleans,
  Louisiana. Association for Computational Linguistics.

\bibitem[{Lewis et~al.(2020)Lewis, Perez, Piktus, Petroni, Karpukhin, Goyal,
  K{\"u}ttler, Lewis, Yih, Rockt{\"a}schel et~al.}]{lewis2020retrieval}
Patrick Lewis, Ethan Perez, Aleksandra Piktus, Fabio Petroni, Vladimir
  Karpukhin, Naman Goyal, Heinrich K{\"u}ttler, Mike Lewis, Wen-tau Yih, Tim
  Rockt{\"a}schel, et~al. 2020.
\newblock Retrieval-augmented generation for knowledge-intensive nlp tasks.
\newblock \emph{Advances in Neural Information Processing Systems},
  33:9459--9474.

\bibitem[{Li et~al.(2024)Li, Peng, Galley, Gao, and Zhang}]{li-etal-2024-self}
Miaoran Li, Baolin Peng, Michel Galley, Jianfeng Gao, and Zhu Zhang. 2024.
\newblock \href {https://doi.org/10.18653/v1/2024.findings-naacl.12}
  {Self-checker: Plug-and-play modules for fact-checking with large language
  models}.
\newblock In \emph{Findings of the Association for Computational Linguistics:
  NAACL 2024}, pages 163--181, Mexico City, Mexico. Association for
  Computational Linguistics.

\bibitem[{Liu et~al.(2024{\natexlab{a}})Liu, Li, Wu, and Lee}]{liu2024visual}
Haotian Liu, Chunyuan Li, Qingyang Wu, and Yong~Jae Lee. 2024{\natexlab{a}}.
\newblock Visual instruction tuning.
\newblock \emph{Advances in neural information processing systems}, 36.

\bibitem[{Liu et~al.(2024{\natexlab{b}})Liu, Zhang, Li, Yan, Gao, Chen, Yuan,
  Huang, Sun, Gao et~al.}]{liu2024sora}
Yixin Liu, Kai Zhang, Yuan Li, Zhiling Yan, Chujie Gao, Ruoxi Chen, Zhengqing
  Yuan, Yue Huang, Hanchi Sun, Jianfeng Gao, et~al. 2024{\natexlab{b}}.
\newblock Sora: A review on background, technology, limitations, and
  opportunities of large vision models.
\newblock \emph{arXiv preprint arXiv:2402.17177}.

\bibitem[{Lu et~al.(2022)Lu, Bartolo, Moore, Riedel, and
  Stenetorp}]{lu-etal-2022-fantastically}
Yao Lu, Max Bartolo, Alastair Moore, Sebastian Riedel, and Pontus Stenetorp.
  2022.
\newblock \href {https://doi.org/10.18653/v1/2022.acl-long.556} {Fantastically
  ordered prompts and where to find them: Overcoming few-shot prompt order
  sensitivity}.
\newblock In \emph{Proceedings of the 60th Annual Meeting of the Association
  for Computational Linguistics (Volume 1: Long Papers)}, pages 8086--8098,
  Dublin, Ireland. Association for Computational Linguistics.

\bibitem[{Luo et~al.(2021)Luo, Darrell, and
  Rohrbach}]{luo-etal-2021-newsclippings}
Grace Luo, Trevor Darrell, and Anna Rohrbach. 2021.
\newblock \href {https://doi.org/10.18653/v1/2021.emnlp-main.545}
  {{N}ews{CLIP}pings: {A}utomatic {G}eneration of {O}ut-of-{C}ontext
  {M}ultimodal {M}edia}.
\newblock In \emph{Proceedings of the 2021 Conference on Empirical Methods in
  Natural Language Processing}, pages 6801--6817, Online and Punta Cana,
  Dominican Republic. Association for Computational Linguistics.

\bibitem[{Mavi et~al.(2022)Mavi, Jangra, and Jatowt}]{mavi2022survey}
Vaibhav Mavi, Anubhav Jangra, and Adam Jatowt. 2022.
\newblock A survey on multi-hop question answering and generation.
\newblock \emph{arXiv preprint arXiv:2204.09140}.

\bibitem[{Mishra et~al.(2022)Mishra, Suryavardan, Bhaskar, Chopra, Reganti,
  Patwa, Das, Chakraborty, Sheth, Ekbal et~al.}]{mishra2022factify}
Shreyash Mishra, S~Suryavardan, Amrit Bhaskar, Parul Chopra, Aishwarya~N
  Reganti, Parth Patwa, Amitava Das, Tanmoy Chakraborty, Amit~P Sheth, Asif
  Ekbal, et~al. 2022.
\newblock Factify: A multi-modal fact verification dataset.
\newblock In \emph{DE-FACTIFY@ AAAI}.

\bibitem[{Ouyang et~al.(2022)Ouyang, Wu, Jiang, Almeida, Wainwright, Mishkin,
  Zhang, Agarwal, Slama, Ray et~al.}]{ouyang2022training}
Long Ouyang, Jeffrey Wu, Xu~Jiang, Diogo Almeida, Carroll Wainwright, Pamela
  Mishkin, Chong Zhang, Sandhini Agarwal, Katarina Slama, Alex Ray, et~al.
  2022.
\newblock Training language models to follow instructions with human feedback.
\newblock \emph{Advances in neural information processing systems},
  35:27730--27744.

\bibitem[{Pace et~al.(2024)Pace, Mallinson, Malmi, Krause, and
  Severyn}]{pace2024west}
Aliz{\'e}e Pace, Jonathan Mallinson, Eric Malmi, Sebastian Krause, and Aliaksei
  Severyn. 2024.
\newblock West-of-n: Synthetic preference generation for improved reward
  modeling.
\newblock \emph{arXiv preprint arXiv:2401.12086}.

\bibitem[{Pan et~al.(2023{\natexlab{a}})Pan, Saxon, Xu, Nathani, Wang, and
  Wang}]{pan2023automatically}
Liangming Pan, Michael Saxon, Wenda Xu, Deepak Nathani, Xinyi Wang, and
  William~Yang Wang. 2023{\natexlab{a}}.
\newblock Automatically correcting large language models: Surveying the
  landscape of diverse self-correction strategies.
\newblock \emph{arXiv preprint arXiv:2308.03188}.

\bibitem[{Pan et~al.(2023{\natexlab{b}})Pan, Wu, Lu, Luu, Wang, Kan, and
  Nakov}]{pan-etal-2023-fact}
Liangming Pan, Xiaobao Wu, Xinyuan Lu, Anh~Tuan Luu, William~Yang Wang, Min-Yen
  Kan, and Preslav Nakov. 2023{\natexlab{b}}.
\newblock \href {https://doi.org/10.18653/v1/2023.acl-long.386} {Fact-checking
  complex claims with program-guided reasoning}.
\newblock In \emph{Proceedings of the 61st Annual Meeting of the Association
  for Computational Linguistics (Volume 1: Long Papers)}, pages 6981--7004,
  Toronto, Canada. Association for Computational Linguistics.

\bibitem[{Pan et~al.(2023{\natexlab{c}})Pan, Pan, Chen, Nakov, Kan, and
  Wang}]{pan-etal-2023-risk}
Yikang Pan, Liangming Pan, Wenhu Chen, Preslav Nakov, Min-Yen Kan, and William
  Wang. 2023{\natexlab{c}}.
\newblock \href {https://doi.org/10.18653/v1/2023.findings-emnlp.97} {On the
  risk of misinformation pollution with large language models}.
\newblock In \emph{Findings of the Association for Computational Linguistics:
  EMNLP 2023}, pages 1389--1403, Singapore. Association for Computational
  Linguistics.

\bibitem[{Press et~al.(2023)Press, Zhang, Min, Schmidt, Smith, and
  Lewis}]{press-etal-2023-measuring}
Ofir Press, Muru Zhang, Sewon Min, Ludwig Schmidt, Noah Smith, and Mike Lewis.
  2023.
\newblock \href {https://doi.org/10.18653/v1/2023.findings-emnlp.378}
  {Measuring and narrowing the compositionality gap in language models}.
\newblock In \emph{Findings of the Association for Computational Linguistics:
  EMNLP 2023}, pages 5687--5711, Singapore. Association for Computational
  Linguistics.

\bibitem[{Ramesh et~al.(2021)Ramesh, Pavlov, Goh, Gray, Voss, Radford, Chen,
  and Sutskever}]{ramesh2021zero}
Aditya Ramesh, Mikhail Pavlov, Gabriel Goh, Scott Gray, Chelsea Voss, Alec
  Radford, Mark Chen, and Ilya Sutskever. 2021.
\newblock Zero-shot text-to-image generation.
\newblock In \emph{International conference on machine learning}, pages
  8821--8831. Pmlr.

\bibitem[{Rombach et~al.(2022)Rombach, Blattmann, Lorenz, Esser, and
  Ommer}]{rombach2022high}
Robin Rombach, Andreas Blattmann, Dominik Lorenz, Patrick Esser, and Bj{\"o}rn
  Ommer. 2022.
\newblock High-resolution image synthesis with latent diffusion models.
\newblock In \emph{Proceedings of the IEEE/CVF conference on computer vision
  and pattern recognition}, pages 10684--10695.

\bibitem[{Sclar et~al.(2023)Sclar, Choi, Tsvetkov, and
  Suhr}]{sclar2023quantifying}
Melanie Sclar, Yejin Choi, Yulia Tsvetkov, and Alane Suhr. 2023.
\newblock Quantifying language models' sensitivity to spurious features in
  prompt design or: How i learned to start worrying about prompt formatting.
\newblock \emph{arXiv preprint arXiv:2310.11324}.

\bibitem[{Shu et~al.(2020)Shu, Mahudeswaran, Wang, Lee, and
  Liu}]{shu2020fakenewsnet}
Kai Shu, Deepak Mahudeswaran, Suhang Wang, Dongwon Lee, and Huan Liu. 2020.
\newblock Fakenewsnet: A data repository with news content, social context, and
  spatiotemporal information for studying fake news on social media.
\newblock \emph{Big data}, 8(3):171--188.

\bibitem[{Talmor and Berant(2018)}]{talmor2018web}
Alon Talmor and Jonathan Berant. 2018.
\newblock The web as a knowledge-base for answering complex questions.
\newblock \emph{arXiv preprint arXiv:1803.06643}.

\bibitem[{Talmor et~al.(2021)Talmor, Yoran, Catav, Lahav, Wang, Asai, Ilharco,
  Hajishirzi, and Berant}]{talmor2021multimodalqa}
Alon Talmor, Ori Yoran, Amnon Catav, Dan Lahav, Yizhong Wang, Akari Asai,
  Gabriel Ilharco, Hannaneh Hajishirzi, and Jonathan Berant. 2021.
\newblock Multimodalqa: Complex question answering over text, tables and
  images.
\newblock \emph{arXiv preprint arXiv:2104.06039}.

\bibitem[{Tan et~al.(2024)Tan, Beigi, Wang, Guo, Bhattacharjee, Jiang, Karami,
  Li, Cheng, and Liu}]{tan2024large}
Zhen Tan, Alimohammad Beigi, Song Wang, Ruocheng Guo, Amrita Bhattacharjee,
  Bohan Jiang, Mansooreh Karami, Jundong Li, Lu~Cheng, and Huan Liu. 2024.
\newblock Large language models for data annotation: A survey.
\newblock \emph{arXiv preprint arXiv:2402.13446}.

\bibitem[{Team et~al.(2023)Team, Anil, Borgeaud, Wu, Alayrac, Yu, Soricut,
  Schalkwyk, Dai, Hauth et~al.}]{team2023gemini}
Gemini Team, Rohan Anil, Sebastian Borgeaud, Yonghui Wu, Jean-Baptiste Alayrac,
  Jiahui Yu, Radu Soricut, Johan Schalkwyk, Andrew~M Dai, Anja Hauth, et~al.
  2023.
\newblock Gemini: a family of highly capable multimodal models.
\newblock \emph{arXiv preprint arXiv:2312.11805}.

\bibitem[{Thorne and Vlachos(2018)}]{thorne-vlachos-2018-automated}
James Thorne and Andreas Vlachos. 2018.
\newblock \href {https://aclanthology.org/C18-1283} {Automated fact checking:
  Task formulations, methods and future directions}.
\newblock In \emph{Proceedings of the 27th International Conference on
  Computational Linguistics}, pages 3346--3359, Santa Fe, New Mexico, USA.
  Association for Computational Linguistics.

\bibitem[{Thorne et~al.(2018)Thorne, Vlachos, Christodoulopoulos, and
  Mittal}]{thorne-etal-2018-fever}
James Thorne, Andreas Vlachos, Christos Christodoulopoulos, and Arpit Mittal.
  2018.
\newblock \href {https://doi.org/10.18653/v1/N18-1074} {{FEVER}: a large-scale
  dataset for fact extraction and {VER}ification}.
\newblock In \emph{Proceedings of the 2018 Conference of the North {A}merican
  Chapter of the Association for Computational Linguistics: Human Language
  Technologies, Volume 1 (Long Papers)}, pages 809--819, New Orleans,
  Louisiana. Association for Computational Linguistics.

\bibitem[{Wang and Shu(2023)}]{wang-shu-2023-explainable}
Haoran Wang and Kai Shu. 2023.
\newblock \href {https://doi.org/10.18653/v1/2023.findings-emnlp.416}
  {Explainable claim verification via knowledge-grounded reasoning with large
  language models}.
\newblock In \emph{Findings of the Association for Computational Linguistics:
  EMNLP 2023}, pages 6288--6304, Singapore. Association for Computational
  Linguistics.

\bibitem[{Wang and Shu(2024)}]{10.1145/3627673.3679821}
Haoran Wang and Kai Shu. 2024.
\newblock \href {https://doi.org/10.1145/3627673.3679821} {Trojan activation
  attack: Red-teaming large language models using steering vectors for
  safety-alignment}.
\newblock In \emph{Proceedings of the 33rd ACM International Conference on
  Information and Knowledge Management}, CIKM '24, page 2347–2357, New York,
  NY, USA. Association for Computing Machinery.

\bibitem[{Wang et~al.(2023)Wang, Zhou, and Sachan}]{wang-etal-2023-lets}
Ruida Wang, Wangchunshu Zhou, and Mrinmaya Sachan. 2023.
\newblock \href {https://doi.org/10.18653/v1/2023.findings-emnlp.791} {Let{'}s
  synthesize step by step: Iterative dataset synthesis with large language
  models by extrapolating errors from small models}.
\newblock In \emph{Findings of the Association for Computational Linguistics:
  EMNLP 2023}, pages 11817--11831, Singapore. Association for Computational
  Linguistics.

\bibitem[{Wang(2017)}]{wang-2017-liar}
William~Yang Wang. 2017.
\newblock \href {https://doi.org/10.18653/v1/P17-2067} {{``}liar, liar pants on
  fire{''}: A new benchmark dataset for fake news detection}.
\newblock In \emph{Proceedings of the 55th Annual Meeting of the Association
  for Computational Linguistics (Volume 2: Short Papers)}, pages 422--426,
  Vancouver, Canada. Association for Computational Linguistics.

\bibitem[{Wei et~al.(2022)Wei, Wang, Schuurmans, Bosma, Xia, Chi, Le, Zhou
  et~al.}]{wei2022chain}
Jason Wei, Xuezhi Wang, Dale Schuurmans, Maarten Bosma, Fei Xia, Ed~Chi, Quoc~V
  Le, Denny Zhou, et~al. 2022.
\newblock Chain-of-thought prompting elicits reasoning in large language
  models.
\newblock \emph{Advances in neural information processing systems},
  35:24824--24837.

\bibitem[{Welbl et~al.(2018)Welbl, Stenetorp, and
  Riedel}]{welbl2018constructing}
Johannes Welbl, Pontus Stenetorp, and Sebastian Riedel. 2018.
\newblock Constructing datasets for multi-hop reading comprehension across
  documents.
\newblock \emph{Transactions of the Association for Computational Linguistics},
  6:287--302.

\bibitem[{Wu et~al.(2024)Wu, Huang, Gao, Chen, Zhang, Wan, Zhou, Zhang, Gao,
  Xiao et~al.}]{wu2024unigen}
Siyuan Wu, Yue Huang, Chujie Gao, Dongping Chen, Qihui Zhang, Yao Wan, Tianyi
  Zhou, Xiangliang Zhang, Jianfeng Gao, Chaowei Xiao, et~al. 2024.
\newblock Unigen: A unified framework for textual dataset generation using
  large language models.
\newblock \emph{arXiv preprint arXiv:2406.18966}.

\bibitem[{Yang et~al.(2022)Yang, Wang, Jin, Li, Lian, and
  Xie}]{yang2022reinforcement}
Ruichao Yang, Xiting Wang, Yiqiao Jin, Chaozhuo Li, Jianxun Lian, and Xing Xie.
  2022.
\newblock Reinforcement subgraph reasoning for fake news detection.
\newblock In \emph{Proceedings of the 28th ACM SIGKDD Conference on Knowledge
  Discovery and Data Mining}, pages 2253--2262.

\bibitem[{Yang et~al.(2018)Yang, Qi, Zhang, Bengio, Cohen, Salakhutdinov, and
  Manning}]{yang-etal-2018-hotpotqa}
Zhilin Yang, Peng Qi, Saizheng Zhang, Yoshua Bengio, William Cohen, Ruslan
  Salakhutdinov, and Christopher~D. Manning. 2018.
\newblock \href {https://doi.org/10.18653/v1/D18-1259} {{H}otpot{QA}: A dataset
  for diverse, explainable multi-hop question answering}.
\newblock In \emph{Proceedings of the 2018 Conference on Empirical Methods in
  Natural Language Processing}, pages 2369--2380, Brussels, Belgium.
  Association for Computational Linguistics.

\bibitem[{Yao et~al.(2023)Yao, Shah, Sun, Cho, and Huang}]{yao2023end}
Barry~Menglong Yao, Aditya Shah, Lichao Sun, Jin-Hee Cho, and Lifu Huang. 2023.
\newblock End-to-end multimodal fact-checking and explanation generation: A
  challenging dataset and models.
\newblock In \emph{Proceedings of the 46th International ACM SIGIR Conference
  on Research and Development in Information Retrieval}, pages 2733--2743.

\bibitem[{Zhang et~al.(2024)Zhang, Gao, Chen, Huang, Huang, Sun, Zhang, Li, Fu,
  Wan et~al.}]{zhang2024llm}
Qihui Zhang, Chujie Gao, Dongping Chen, Yue Huang, Yixin Huang, Zhenyang Sun,
  Shilin Zhang, Weiye Li, Zhengyan Fu, Yao Wan, et~al. 2024.
\newblock Llm-as-a-coauthor: Can mixed human-written and machine-generated text
  be detected?
\newblock In \emph{Findings of the Association for Computational Linguistics:
  NAACL 2024}, pages 409--436.

\bibitem[{Zlatkova et~al.(2019)Zlatkova, Nakov, and
  Koychev}]{zlatkova-etal-2019-fact}
Dimitrina Zlatkova, Preslav Nakov, and Ivan Koychev. 2019.
\newblock \href {https://doi.org/10.18653/v1/D19-1216} {Fact-checking meets
  fauxtography: Verifying claims about images}.
\newblock In \emph{Proceedings of the 2019 Conference on Empirical Methods in
  Natural Language Processing and the 9th International Joint Conference on
  Natural Language Processing (EMNLP-IJCNLP)}, pages 2099--2108, Hong Kong,
  China. Association for Computational Linguistics.

\end{thebibliography}

\appendix

\clearpage
\section{Appendix}
\label{sec:appendix}

\subsection{Dataset Example}
\label{sec:dataset_example}
Here is an example of dataset schema from MMCV:

\begin{tcolorbox}[example]
\footnotesize{\texttt{\textbf{claim}: Stoke City, a club that was part of the top-tier league before 1992, was promoted to the highest level of English football in 2018. \\
\textbf{wiki\_context}: The Premier League is the highest level of the English football league system. Contested by 20 clubs, it operates on a system of promotion and relegation with the English Football League (EFL). Seasons usually run from August to May, with each team playing 38 matches: two against each other, one home and one away. Most games are played on weekend afternoons, with occasional weekday evening fixtures.\\
\textbf{text\_evidence}: [
"f369cee1ca92368c8b1ea564c5e41fc1"
]\\
\textbf{image\_evidence}: []\\
\textbf{table\_evidence}: [
  "c120efadd518b5f32c11d40b456c8570"
]\\
\textbf{label}: SUPPORT
}}
\end{tcolorbox}

\noindent Additional examples of 1-hop, 2-hop, 3-hop, and 4-hop claims are listed in Table \ref{tab:dataset_example}

\subsection{Experiment Prompt}
\label{sec:experiment_prompt}
\noindent \textbf{Claim Verification Prompt.}
To test MLLMs' claim verification performance under zero-shot settings, we follow \cite{geng2024multimodal} and use the following prompt.
\begin{tcolorbox}[prompt]
\footnotesize{\texttt{Given a claim and evidence  (which can be text, table, or an image), determine whether the claim is SUPPORT or REFUTE by the evidence. \\
\\
Use the following format to provide your answer:\\
Prediction: [True or False]\\
Explanation: [put your evidence and step-by-step reasoning here]\\
Confidence Level: [please show the percentage]\\
\\
Note: The confidence level indicates the degree of certainty you have about your answer and is represented as a percentage. For instance, if your confidence level is 80\%, it means you are 80\% certain that your answer is correct and there is a 20\% chance that it may be incorrect.
}}
\end{tcolorbox}

\noindent \textbf{Claim Generation Prompt.}
We use the following prompt to convert multimodal QA pairs into claim candidates:
\begin{tcolorbox}[prompt]
\footnotesize{\texttt{You are an expert in converting question-answers into claims.\\
For example: Question: Telos was an album by a band who formed in what city? Answer: Indianapolis.\\
Claim: Telos was an album by a band formed in Indianapolis.\\
\\
Convert the question-answer into claim. Return only the claim and nothing else.
}}
\end{tcolorbox}

\noindent \textbf{Claim Modification Prompt.}
We use the following prompt to modify the claim candidates:
\begin{tcolorbox}[prompt]
\footnotesize{\texttt{Generate a multi-hop specific claim based on the given general claim and Wikipedia context. The specific claim should:\\
\\
Incorporate information from Wikipedia context.\\
Provided context should always be factually correct.\\
Obscure key information by:\\
    a) Replacing one or two central entities with related fact using the Wikipedia context.\\
    b) Alluding to critical details without explicitly stating them. Claim should be short and concise. \\
For example:\\
\\
- General Claim: The Mona Lisa is a famous painting by Leonardo da Vinci.\\
- Wikipedia Context: The Mona Lisa is a half-length portrait painting by Italian artist Leonardo da Vinci. Considered an archetypal masterpiece of the Italian Renaissance, it has been described as \"the best known, the most visited, the most written about, the most sung about, the most parodied work of art in the world\". The painting's novel qualities include the subject's enigmatic expression, the monumentality of the composition, the subtle modelling of forms, and the atmospheric illusionism. It is housed in the Louvre Museum in Paris, where it was first put on display in 1797.\\
- Specific Claim: The Mona Lisa is a half-length portrait painting created by Italian artist who is considered as archetypal masterpiece of the Italian Renaissance.
}}
\end{tcolorbox}

\noindent \textbf{Claim Refinement Prompt.}
We use the following prompt to refine the claim candidates:
\begin{tcolorbox}[prompt]
\footnotesize{\texttt{You are tasked with improving a claim focusing on three key areas: Fluency, Correctness, and Clearness. Your goal is to enhance the text while maintaining its original meaning and intent.\\
\\
Improvement Criteria:\\
Fluency:\\
1. Review the text for grammar, syntax, and punctuation errors.\\
2. Rephrase any awkward or unnatural sentences to make the text flow more smoothly.\\
3. Ensure that the text reads naturally and is easy to follow.\\
\\
Correctness:\\
1. Verify the factual accuracy of the content and correct any errors.\\
2. Ensure that the text adheres to the prompt's instructions.\\
3. Clarify any ambiguities and correct any inconsistencies in the information presented.\\
\\
Clearness:\\
1. Simplify complex sentences or ideas to make the text easier to understand.\\
2. Improve the organization of ideas to enhance readability.\\
3. Ensure that the message is conveyed clearly and effectively, eliminating any confusion or ambiguity.\\
\\
Final Output:\\
Once you have made the necessary improvements, provide the revised text. Ensure that the improved version is more fluent, accurate, and clear than the original while preserving the original meaning and intent.\\
Example Improvement:\\
Original Claim: "The results of the survey was very positive, with many respondents saying that they would recommend the service to others, however, some were also mentioned issues with the customer support."\\
Improved Claim: "The survey results were overwhelmingly positive, with many respondents stating they would recommend the service to others. However, some also noted issues with customer support.
}}
\end{tcolorbox}

\newpage
\subsection{Annotation Guidelines}
\label{sec:annotation_guidelines}
We ask our annotators to score the quality of the claim from three aspects: fluency, correctness, and clearness. Here is the detailed guidelines provided to the human annotators.

\begin{tcolorbox}[guideline]
    $\triangleright$ \textbf{Scoring Criteria:}\\
    Fluency: Rate on a scale of 1-4. \\
    Correctness: Rate on a scale of 1-3. \\
    Clearness Rate on a scale of 1-3. \\

    $\triangleright$ \textbf{Fluency (1-4):}\\
    4: Excellent - Reads naturally, no awkward phrasing. \\
    3: Good - Mostly smooth, minor phrasing issues. \\
    2: Fair - Several awkward phrases or constructions. \\
    1: Poor - Difficult to read, very unnatural phrasing. \\

    $\triangleright$ \textbf{Correctness (1-3):}\\
    3: Fully correct - All information is accurate. \\
    2: Partially correct - Some information is accurate, some errors. \\
    1: Incorrect - Significant factual errors or misrepresentations. \\

    $\triangleright$ \textbf{Clearness (1-3):}\\
    3: Very clear - Easy to understand, no ambiguity. \\
    2: Somewhat clear - Some parts may be confusing or ambiguous. \\
    1: Unclear - Difficult to understand the intended meaning. \\
\end{tcolorbox}

\subsection{Crowd Worker Interface}
\label{sec:ui}
\begin{figure*}[ht]
    \centering
    \includegraphics[width=0.99\linewidth]{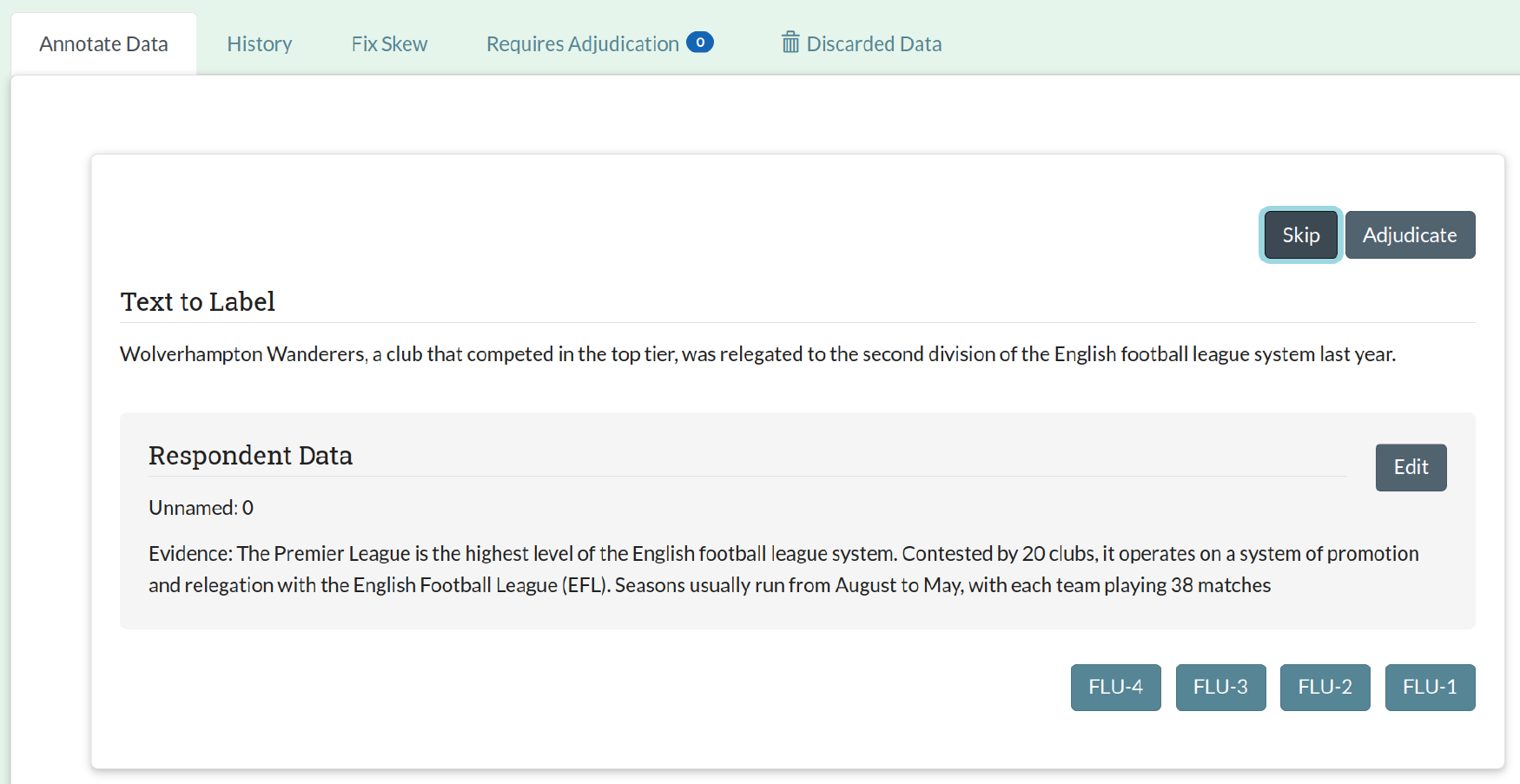}
    \caption{UI for human annotators.}
    \label{fig:ui}
\end{figure*}

We use SMART \cite{chew2019smart}, an open-source project designed to help data scientists and research teams efficiently build labeled training datasets for supervised machine learning tasks. Figure \ref{fig:ui} shows an example of the worker interface during scoring procedure.

\begin{table*}[ht]
    \centering
    \renewcommand{\arraystretch}{1.2}%
    \resizebox{0.99\linewidth}{!}{%
    \begin{tabular}{p{0.02\linewidth}p{0.4\linewidth}p{0.58\linewidth}}
    \toprule
    \textbf{\#H} & \textbf{Claim} & \textbf{Evidence}  \\ \midrule
    1 & \textbf{Claim:} Marisa Coughlan played the role of Chante Lefort on television in 1996.   & \raisebox{-0.5\height}{\includegraphics[width=\linewidth]{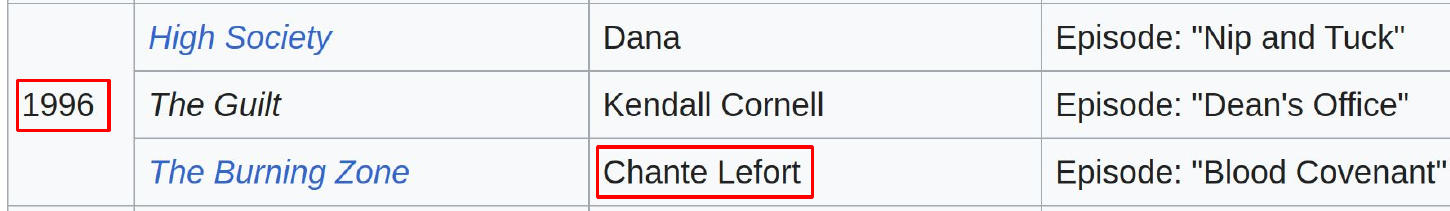}} \\[2ex]
    \hline
    2 & \textbf{Claim:} The driver seen signing autographs outside had a significant points total during a specific race in 2001 while competing for a well-known team in stock car racing.  & 
    \raisebox{-0.7\height}{%
    \begin{minipage}[t]{\linewidth}
    \vspace{0.1cm}
        \includegraphics[width=0.27\linewidth]{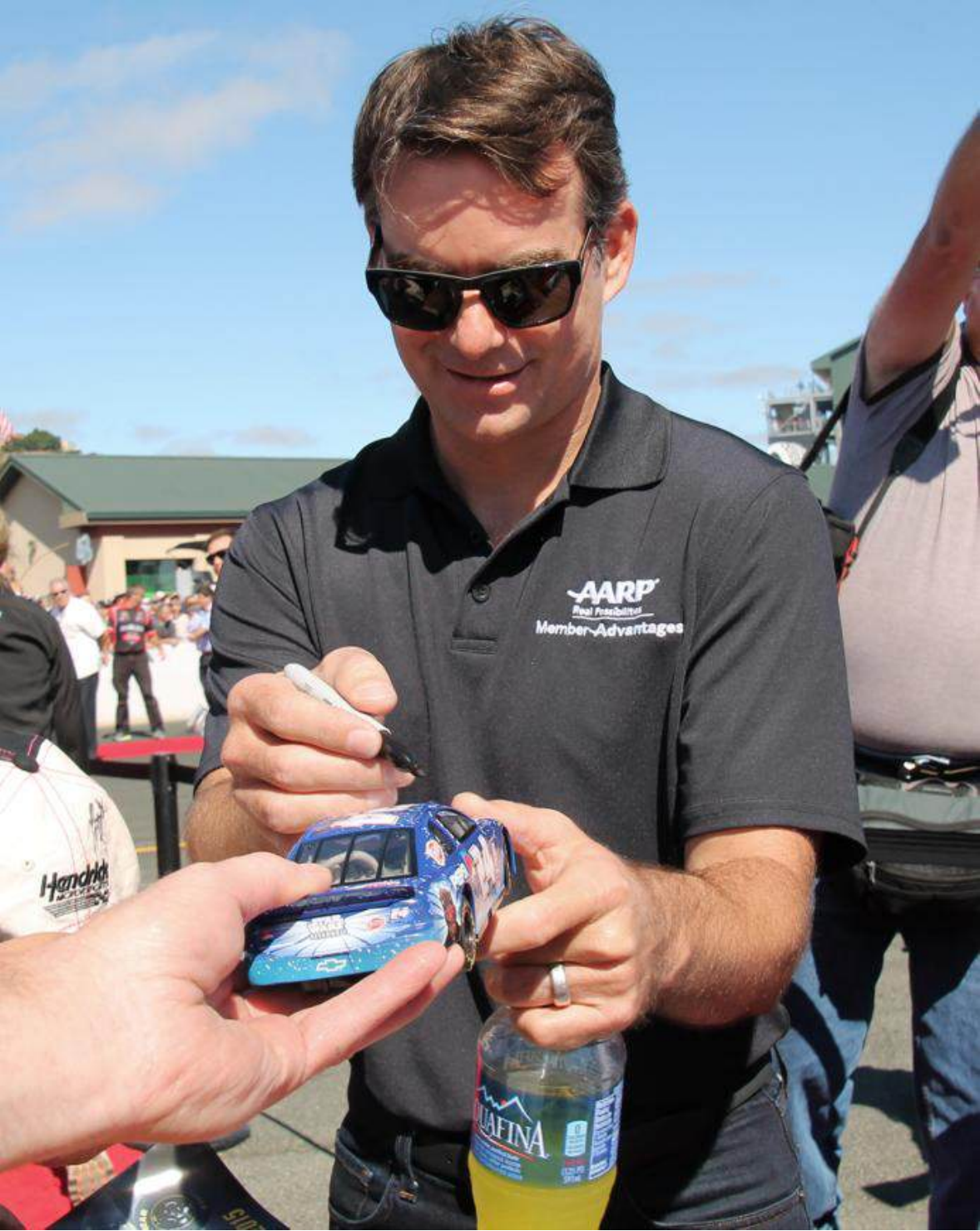}%
        \includegraphics[width=0.73\linewidth]{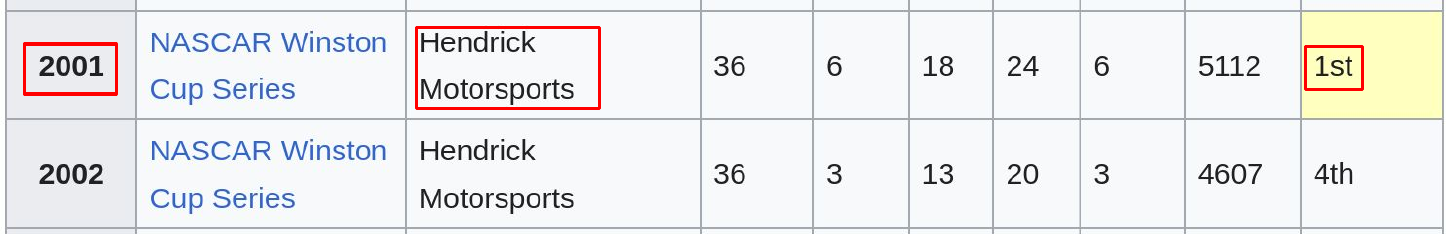}
    \end{minipage}} \\[2ex]
    \hline
    3 & \textbf{Claim:} The Green Bay Packers were one of the two teams that played in the first Super Bowl and also faced the New York Giants at MetLife Stadium during the 2013 regular season. & \textbf{Doc A:} The first AFL-NFL World Championship Game in professional American football, known retroactively as Super Bowl I and ... ... \newline \textbf{Doc B:} The National Football League (NFL) champion Green Bay Packers defeated the American Football League (AFL) champion Kansas City Chiefs ... ... \newline\textbf{Table:} Not Included Here \\[2ex]
    \hline
    4 & \textbf{Claim:} The team that was promoted to the Premier League in the 2018-19 season received a higher accolade in the Third Division PFA Team of the Year during the 1980s than a club renowned for its West London rivalries. & \textbf{Doc A:} Manchester City are the defending champions. Wolverhampton Wanderers, Cardiff City and Fulham join as the promoted clubs from the 2017–18 EFL Championship. ... ... \newline \textbf{Doc B:} ... They will replace West Bromwich Albion, Swansea City and Stoke City who were relegated to the 2018–19 EFL Championship. ... \newline \textbf{Table:} Not Included Here \newline \textbf{Image:} Not Included Here \\[2ex]
    \bottomrule
    \end{tabular}
    }
    \caption{Examples of 1-hop, 2-hop, 3-hop and 4-hop claims from \textsc{MMCV}.}
    \label{tab:dataset_example}
\end{table*}

\begin{figure*}[ht]
    \centering
    \includegraphics[width=0.99\linewidth]{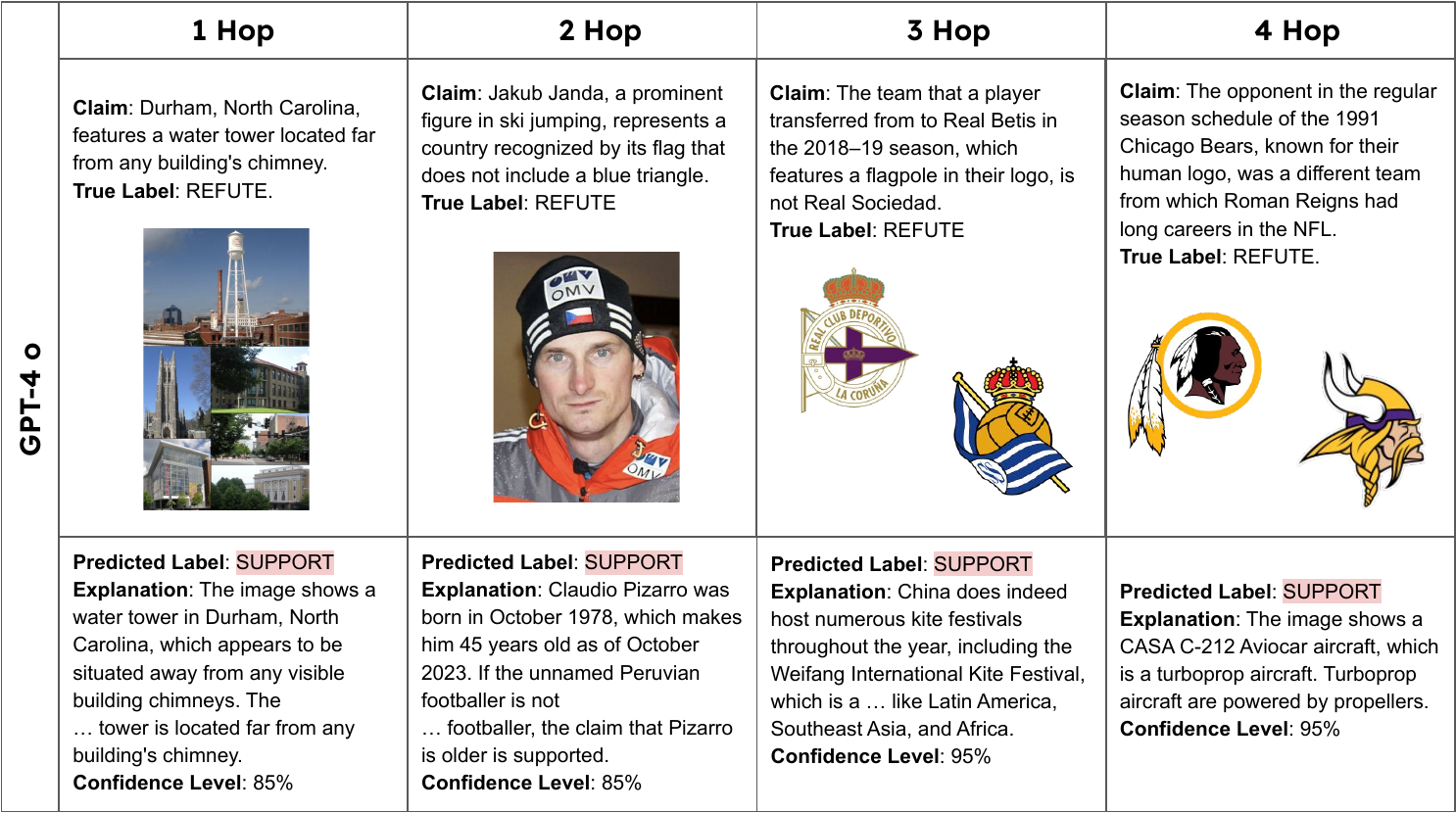}
    \caption{Error Analysis: Gemini}
    \label{fig:gpt4_error}
\end{figure*}

\vspace{-0.5cm}

\begin{figure*}[ht]
    \centering
    \includegraphics[width=0.99\linewidth]{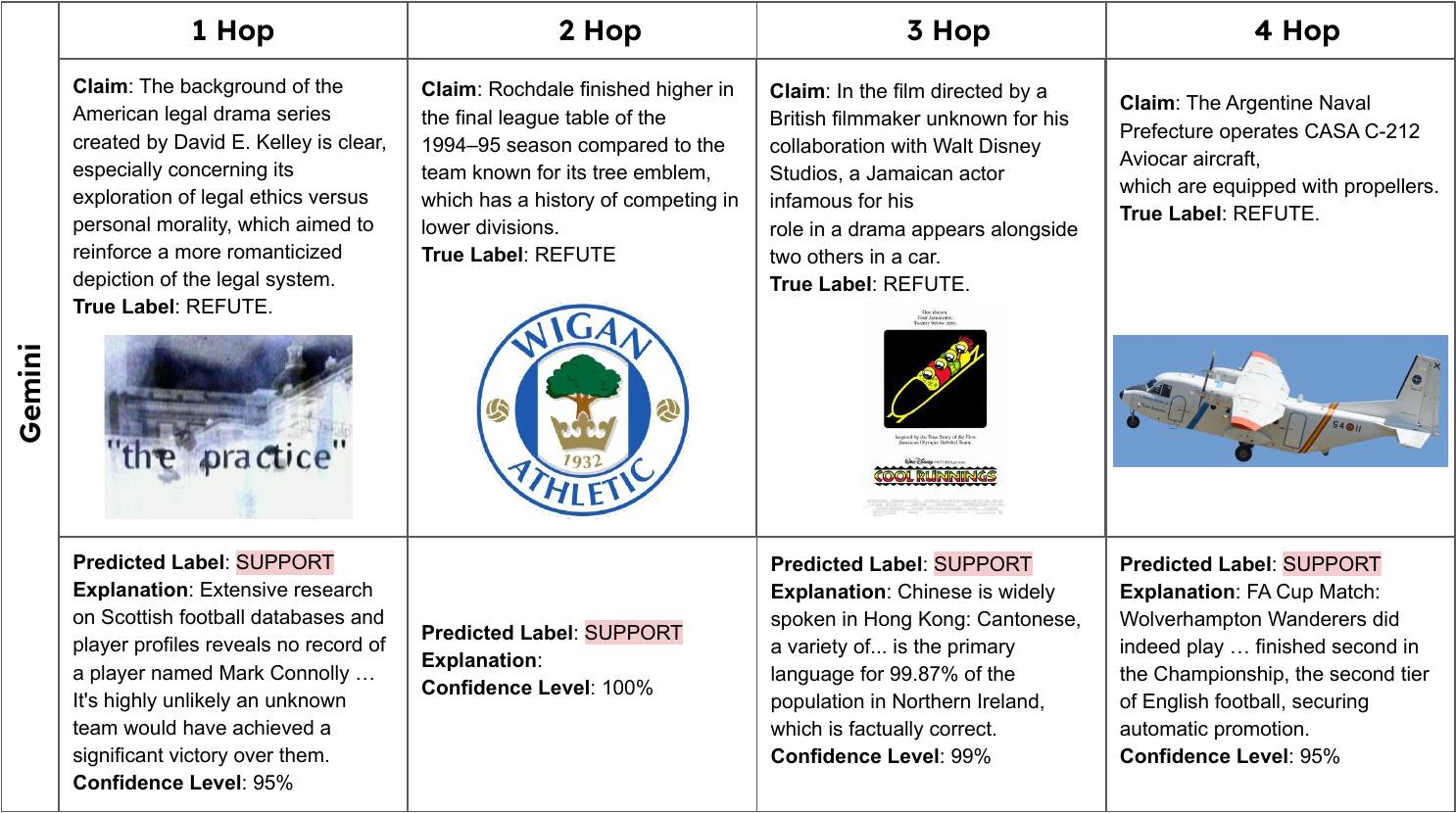}
    \caption{Error Analysis: GPT4-o}
    \label{fig:gemini_error}
\end{figure*}

\begin{figure*}[!t]
    \centering
    \includegraphics[width=0.99\linewidth]{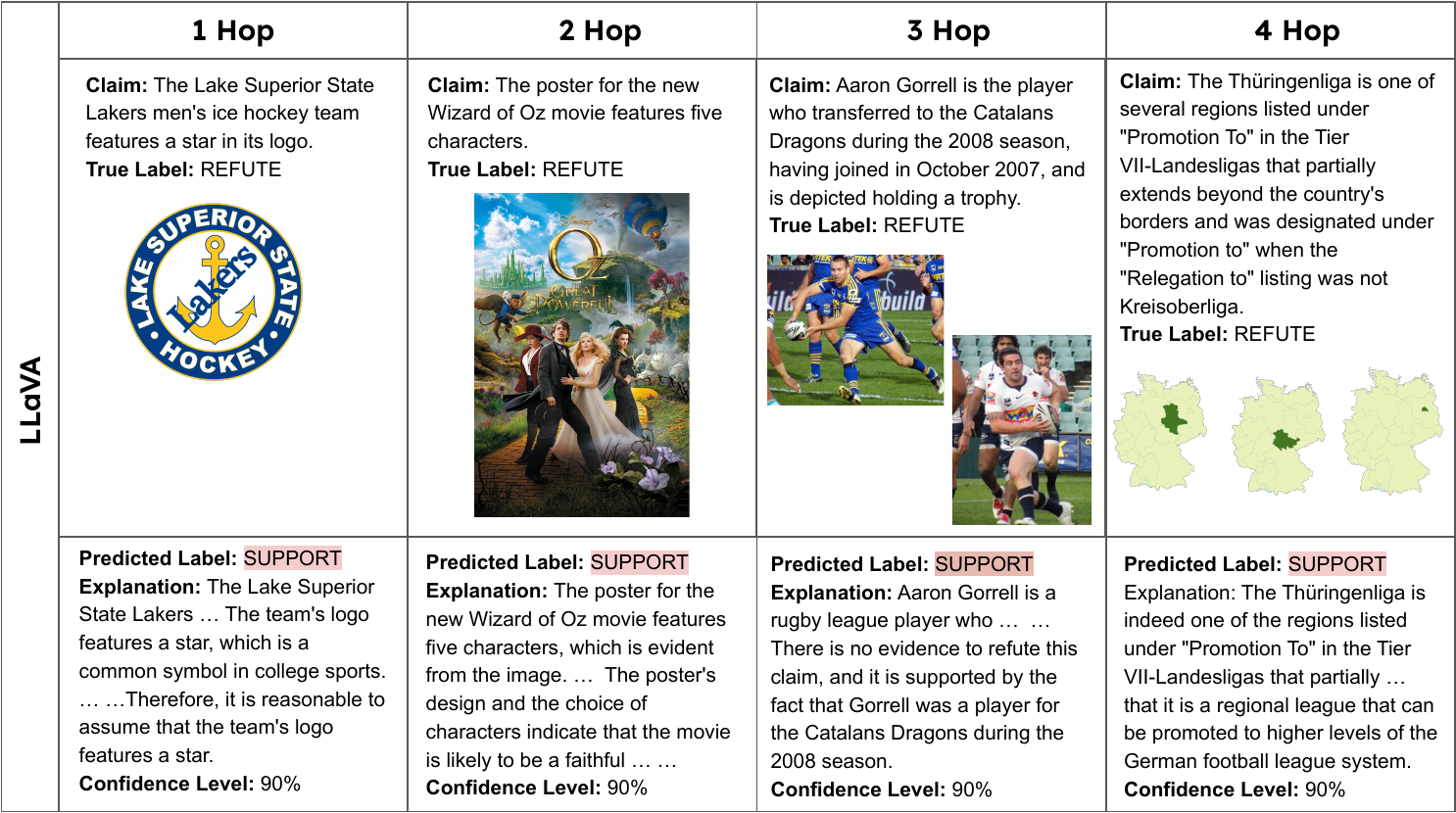}
    \caption{Error Analysis: LLaVA}
    \label{fig:llava_error}
\end{figure*}

\end{document}